\documentclass{article}
\usepackage{jfrExamplee}
\usepackage{graphicx}
\graphicspath{ {Images} }
\usepackage{natbib} 
\usepackage{apalike}
\usepackage{setspace}
\usepackage{epstopdf}
\usepackage{url}
\usepackage{multirow}
\usepackage{hhline}
\usepackage{threeparttable}
\usepackage{caption}
\usepackage{subcaption}

\title{Image Segmentation for Fruit Detection and Yield Estimation in Apple Orchards}

\author{Suchet Bargoti, James P.~Underwood \\
Australian Centre for Field Robotics\\
The University of Sydney\\
NSW, 2006, Australia\\
\texttt{s.bargoti j.underwood @acfr.usyd.edu.au} \\
}

\begin{document}

\maketitle

\begin{abstract}
Ground vehicles equipped with monocular vision systems are a valuable source of high resolution image data for precision agriculture applications in orchards. This paper presents an image processing framework for fruit detection and counting using orchard image data. A general purpose image segmentation approach is used, including two feature learning algorithms; multi-scale Multi-Layered Perceptrons (MLP) and Convolutional Neural Networks (CNN). These networks were extended by including contextual information about how the image data was captured (metadata), which correlates with some of the appearance variations and/or class distributions observed in the data. The pixel-wise fruit segmentation output is processed using the Watershed Segmentation (WS) and Circular Hough Transform (CHT) algorithms to detect and count individual fruits. Experiments were conducted in a commercial apple orchard near Melbourne, Australia. The results show an improvement in fruit segmentation performance with the inclusion of metadata on the previously benchmarked MLP network. We extend this work with CNNs, bringing agrovision closer to the state-of-the-art in computer vision, where although metadata had negligible influence, the best pixel-wise F1-score of $0.791$ was achieved. The WS algorithm produced the best apple detection and counting results, with a detection F1-score of $0.858$. As a final step, image fruit counts were accumulated over multiple rows at the orchard and compared against the post-harvest fruit counts that were obtained from a grading and counting machine. The count estimates using CNN and WS resulted in the best performance for this dataset, with a squared correlation coefficient of $r^2=0.826$. 
\end{abstract}

\section{Introduction}
\label{sec:intro}

Yield estimation and mapping in orchards is important for growers as it facilitates efficient utilisation of resources and improves returns per unit area and time. With accurate knowledge of yield distribution and quantity, a grower can efficiently manage processes such as chemigation, fertigation and thinning. Yield estimation also allows the grower to plan ahead of time their harvest logistics, crop storage and sales. The standard approach to get yield information is currently manual sampling, which is labour intensive, expensive and often destructive. Constrained by these costs, sampling is often done over a few individual crops and the measures are extrapolated over the entire farm. Inherent human sampling bias and sparsity in the measurements can result in inaccurate yield estimation.

Recent advances in robotics and automation enable us to gather large scale data with high spatial and temporal resolution. Unmanned Ground Vehicles (UGVs) or farmer operated machinery can be equipped with standard colour cameras to capture a detailed representation over large farms. Robust and accurate image processing techniques are required to extract high level information from this data such as crop location, health, maturity, crop load (yield), spatial distribution etc. Image segmentation is the process of evaluating a semantic description of an image at a pixel or super-pixel level. For image data captured at an orchard, this means automatically labelling each pixel, or groups of pixels as representing fruits, flowers, trunks, branches and foliage. The parsed information can then be used in higher level tasks such as individual fruit detection, crop health analysis and tree detection/branch modelling. This provides the grower with a rich farm inventory and enables further robotic operations such as autonomous pruning, harvesting, variable rate spraying etc.

A standard approach for image segmentation is to transform image regions into discriminative feature representations and parse them through a trained classifier, assigning each region a specific label. For identifying orchard fruits this means extraction of a feature space that captures properties unique to the fruit such as its colour, texture, reflection, position, shape etc. Farm image data is generally prone to a wide range of intra-class (within class) variations due to illumination conditions, occlusions, clustering, camera view-point, tree types, seasonal maturity (translating to a different fruit size, shape and colour) etc. Therefore, the feature space and the classifier need to be invariant to such characteristics. 

Typically, prior works utilise hand engineered features to encode visual attributes that discriminate fruits from non-fruit regions \citep{Payne2014,Wang2013,Nuske2014}. Although these approaches are well suited for the dataset they are designed for, the feature encoding is generally unique to a specific fruit and the conditions under which the data were captured. As a result, the methods are often not transferable to other crops/datasets. In contrast, supervised feature learning approaches can be used to automatically learn transformations that capture the data distribution, enabling their use with different datasets \citep{Hung2013}. Such approaches have high model complexity and utilise extensive training exemplars to model the variations in the data \citep{Krizhevsky2012}. 

Image classification algorithms have also been shown to benefit from prior knowledge about scene structure. For example, \cite{tighe2013superparsing} and \cite{Brust15:CPN} specify a spatial prior over the labelled classes to aid image segmentation of public image datasets such as LabelMeFacade and SIFT Flow. A greater wealth of prior knowledge is often available in field robotics applications, as we often have access to contextual information about how the data were captured. For a typical image dataset, such contextual information, which we refer to as metadata, can include camera trajectories, vehicle location, type of tree/fruit being scanned, distance from camera to trees, sun position, illumination incident angle, weather conditions, etc. While a direct physical model of how particular meta-parameters affect the data is not available, aspects of the relationship between metadata and object classes can be learnt. Where there is correlation between metadata and appearance variations and/or class distributions, including metadata can improve classification performance. Available at no extra cost to typical data capturing process, our previous works \citep{Bargoti2015IROS,Bargoti2016ICRA} have illustrated the use of metadata in allowing simpler classifiers to capture that space, and provide a performance boost, leading to similar performance with reduced training exemplars. 

\begin{figure}[htb!]
	\centering
		\includegraphics[width=\textwidth]{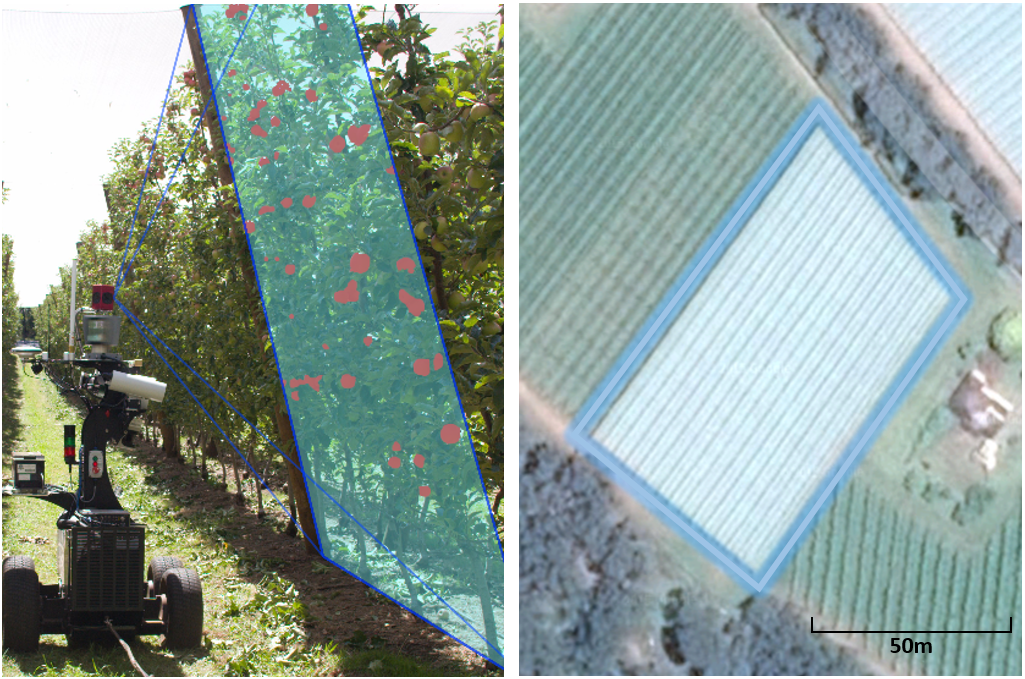}
		\caption{Left: The research ground vehicle Shrimp traversing between rows at an apple orchard, capturing tree image data. Location of apples manually illustrated in the field of view of the camera. Right: Satellite view of the $0.5\ ha$ orchard test block.}
	\label{fig:shrimp-orchard}
\end{figure}

This paper presents a study of different image segmentation frameworks for the task of image fruit segmentation. A conventional visible light camera mounted on a UGV is used to capture images of trees during pre-harvest season as the vehicle drives between different rows at an apple orchard block near Melbourne, Australia (Figure \ref{fig:shrimp-orchard}). We explore multiple supervised feature learning approaches of varying model complexities, while studying the effects of metadata towards image segmentation performance. The impact of accurate image segmentation for agricultural objectives is evaluated, including fruit detection and yield estimation compared to real-world fruit counts. We subsequently provide an orchard block yield map, which can enable the grower to optimise their farm operations. This paper extends from our previous work with evaluation using new segmentation architectures including different configurations of the previously used MLP architecture \citep{Bargoti2016ICRA} and the more widely used CNNs. Additional developments are also made for the fruit detection algorithms and performance metrics, and an in-depth discussion about the practical viability of the image processing approach is presented. The primary contributions from this paper are:

\begin{itemize}
\item Image fruit segmentation analysis using the previously benchmarked multi-scale Multi-Layered Perceptron \citep{Hung2013,Bargoti2016ICRA} and the state-of-the-art Convolutional Neural Networks. 
\item A study of the utility of different metadata sources and their inclusion within the different classification architectures and training configurations. 
\item Analysis of the impact of accurate image segmentation towards fruit detection and yield estimation. 
\end{itemize}

\begin{figure}[b!]
	\centering
		\includegraphics[width=\textwidth]{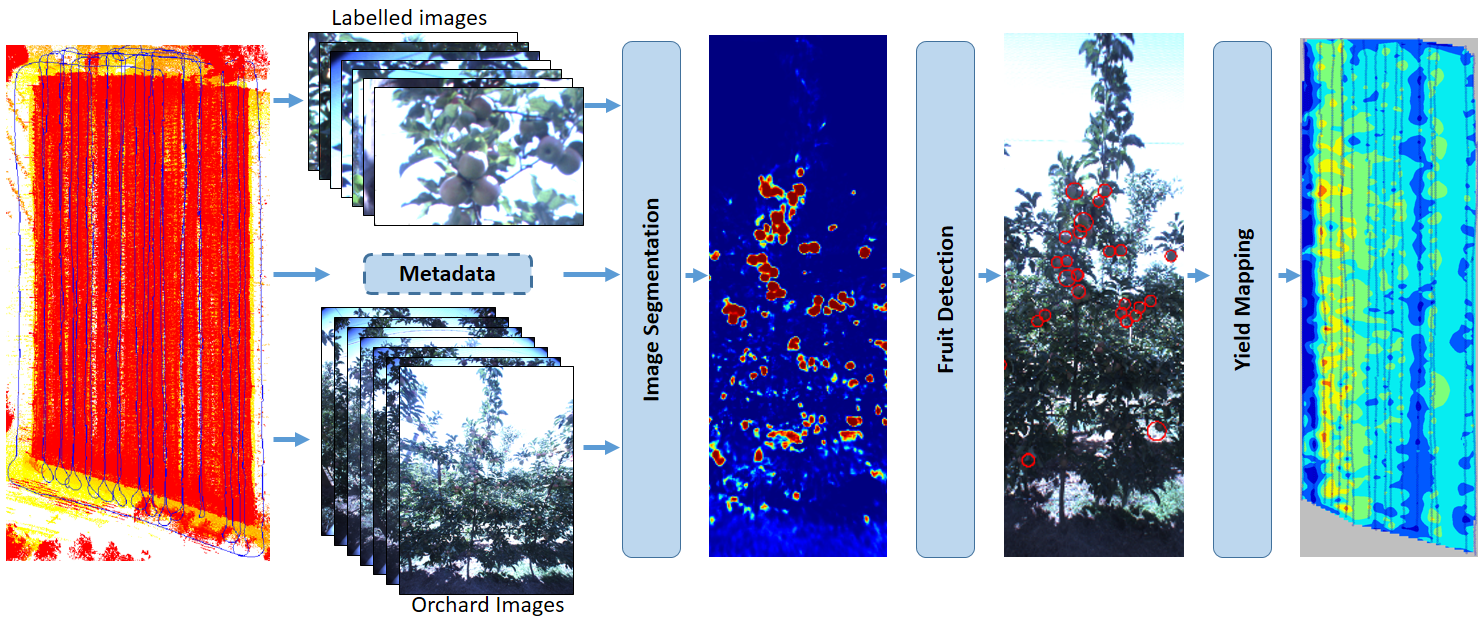}
		\caption{The image segmentation, fruit detection and yield mapping pipeline. An image segmentation model is trained using labelled images extracted sparsely from the farm. This is used to segment the dense image data captured by a UGV. Detection algorithms are applied to the segmentation output to identify individual apples and accumulate the counts over the farm. The result is a dense farm yield map, which can be calibrated with ground truth counts to provide a yield estimate.}
	\label{fig:flow-chart}
\end{figure}

The remainder of this paper is organized as follows. Section \ref{sec:relatedwork} presents related work on image classification in outdoor scenes, both in the agricultural and the general computer vision context. The image processing components are presented over Sections \ref{sec:method} to \ref{sec:detection}, following the computation pipeline illustrated in Figure \ref{fig:flow-chart}. Section \ref{sec:method} presents the different classification frameworks with the inclusion of additional metadata. In Section \ref{sec:setup} we outline the experimental setup and the image dataset, followed by Section \ref{sec:seg-results} presenting the image segmentation results. 
Section \ref{sec:detection} focuses on fruit detection and yield estimation performed over the image segmentation output. We present the practical lessons learnt in Section \ref{sec:discussion}, including new insights into image processing for orchard data. We conclude in Section \ref{sec:conclusion}, discussing the future directions of this work. 

\section{Related Work}
\label{sec:relatedwork}

Computer vision in agriculture, agrovision \citep{Kapach:2012}, has been explored in multiple literature studies for the purposes of fruit detection and yield estimation \citep{Jimenez2000a,Payne2014Book,Kapach:2012}. Agrovision literature is typically data specific, designed for the task at hand, and can often be very heuristic when compared against the most recent work in the general computer vision community \citep{Kapach:2012}. In this section we discuss the key approaches used in agrovision while comparing them against the state-of-the-art techniques used in computer vision.  

Image processing at orchards spans a large variety of fruits such as grapes \citep{Nuske2014,Font2015}, mangoes \citep{Chhabra2012,Payne2014}, apples \citep{Linker2012,Wang2013,Stajnko2009,Silwal2014,Ji2012,Kim2015,Hung2015}, citrus \citep{Li2011,Annamalai2004,Sengupta2014,Regunathan2005,Qiang2014}, kiwifruit \citep{Wijethunga2009} and peaches \citep{Kurtulmus2014}. Fruit classification is generally performed by transforming image regions into discriminative feature spaces and using a trained classifier to associate them to either fruit regions or background objects such as foliage, branches, ground etc.  If conducted densely, image regions are contextual windows neighbouring every pixel in the image and the output is a densely segmented image. Post-processing techniques can then be applied to differentiate individual objects of interest. A detection specific approach on the other hand, reduces the region search space by initially performing key-point detection. Here interesting image regions (possible fruit candidates) are first extracted using manually tuned constraints designed for the particular dataset. This is followed by feature extraction and classification as before. 

Fruit detection through key-point extraction and classification is often applied over vineyards and orchards. For example, \cite{Nuske2014} exploits radial symmetries in the specular reflection of the individual berries to extract key-points, which are then classified as berries or not-berries. Using key-points allows distinct grape-berries to be identified, which is important to extract measurements that are invariant to the stage of the berry development. To detect citrus fruits, \cite{Sengupta2014} first use Circular Hough Transforms (CHT) to extract key-points. Alternatively, \cite{Liu2015} and \cite{Song2014} use simple colour classifiers for key-point extraction for grape bunches and peppers respectively. For fruit detection, image patches are extracted around each key-point and a combination of colour and texture filters are computed. The patches can then be classified as fruits or not-fruits using a trained classifier such as a Support Vector Machine (SVM) or a randomised KD-forest. 

Image segmentation on the other hand returns a rich likelihood map of the fruits, onto which a threshold can be applied to obtain a binary fruit mask detailing regions of the image containing fruit \citep{Payne2014,Linker2012,Yamamoto2014,Sa2015}. \cite{Payne2014} designs a set of heuristic measures based on local colours and textures to classify individual pixels as mangoes or non-mangoes. Blob extraction was done on the resultant binary mask to identify individual mangoes. \cite{Linker2012} incorporates further post-processing for apple detection where individual blobs are expanded, segmented and combined to manage occluded fruits and fruit clusters. \cite{Stajnko2009} instead uses shape analysis and template matching to extract circular apples from the segmented image. \cite{Yamamoto2014} implements a second classification component on tomato blobs extracted via image segmentation to remove any background detections. 

Typically, orchard image data is subject to highly variable illumination conditions, shadowing effects, fruits/crops of different shapes and sizes, captured over different seasons, etc \citep{Payne2014Book}, which makes classification a challenging task. To simplify and minimise the variations in the data, one can enforce constraints on the environment or the data gathering operation. For example, pepper detection in \cite{Song2014} is conducted in a greenhouse with controlled illumination conditions. Equivalently, in \cite{Nuske2014,Payne2014,Font2015} the data is captured at night using strobes, which significantly restricts the illumination variance. However, in orchards it is generally more practical to conduct large scale experiments under natural day-light conditions, and for commercial applications simple hardware such as cameras can be easily incorporated onto tractors, which operate more frequently during the day\footnote{Night time tractor operations are also common for some fruits at some times of the year, but day/night operation will allow most flexibility for adoption.}. Therefore, image classification under natural illumination conditions is an open and important problem.

As stated in the previous section, hand engineered feature encoding often restricts methods to particular fruits/datasets as they are designed to capture data specific inter-class variations (i.e. differences between trees, leaves and fruits), while being invariant to the intra-class variations. Although the methods stated above have produced promising performance over the respective fruits/datasets, they are distinct and ad hoc, seldom replicated, and often disconnected from progress in the general computer vision literature \citep{Kapach:2012}. For widespread application, it would be more efficient to have unified image processing approach compatible with different fruits or capturing configurations. 

A general purpose adaptive feature learning algorithm is therefore desirable, as proposed in \citep{Hung2013}. Here, a pixel-wise image segmentation framework was presented, that utilises a multi-scale feature learning architecture to learn relevant feature transformations for pixel-level classification. The proposed approach was shown to outperform classification using hand engineered features. Additionally, the same architecture has been used for different datasets such as almonds \citep{Hung2013}, apples \citep{Hung2015,Bargoti2015IROS,Bargoti2016ICRA} and tree trunks \citep{Bargoti:2015} without any changes to the image segmentation pipeline. However, even though such a feature learning approach avoids the need for manually designed features, there are opportunities for significant improvements, particularly to address the reliability in adverse illumination conditions, such as underexposed and overexposed images, where performance was observed to deteriorate. 

There have been major developments in the state-of-the-art methods for image segmentation outside the agrovision literature. With the advancements of parallel computing using GPUs, deeper neural network architectures, which host a significantly larger number of model parameters, are showing potential in capturing large variability in data \citep{Krizhevsky2012}. For example, \cite{Ning2005,Ciresan2012,PinheiroC13,GaninL14} use multi-layered Convolutional Neural Networks (CNNs) for image segmentation, where individual patches representing contextual regions around pixels are densely classified in an image. More recently, CNNs have been shown to yield improved segmentation performance when a spatial prior on the classes is available. \cite{Brust15:CPN} perform road image segmentation while incorporating the pixel position, to help the classifier learn that road pixels are pre-dominantly found near the bottom half of images. 

In this paper we evaluate the performance of a previously benchmarked multi-scale Multi-Layered Perceptron (MLP) architecture for image segmentation in agrovision \citep{Hung2013} including an extension with metadata, which we have shown improves performance significantly in \cite{Bargoti2015IROS,Bargoti2016ICRA}. We further extend this study with comparison against state-of-the-art CNNs, with and without the addition of metadata, showing improved pixel classification performance. We evaluate the utility of improved image segmentation towards fruit detection and yield estimation. With this we also shorten the gap between image processing techniques used in agrovision to the current work in computer vision literature, which is a limitation addressed in the literature survey conducted in \cite{Kapach:2012}.

\section{Image Segmentation}
\label{sec:method}

Image segmentation is the task of transforming individual pixels in an image into class labels. In this paper we present multiple image segmentation architectures for the binary classification of orchard image data into fruit/non-fruit classes. These include a multi-scale MLP and a state-of-the-art deep CNN architecture. We then extend these neural network architectures with the inclusion of orchard metadata. All network training is done at the pixel level, however inference is performed over the whole image, resulting in a dense probabilistic output of the fruit and non-fruit classes. This can be used to obtain a binary fruit mask and subsequently perform fruit detection or yield estimation as detailed in Section \ref{sec:detection}. 

\subsection{Multi-scale Multi-Layered Perceptron}

Given the success of the image segmentation framework in \cite{Hung2013} for different fruit types, the reference segmentation architecture in this paper is based on a multi-scale Multi-Layered Perceptron (which we denote as ms-MLP). The classifier takes as input a contextual window around individual pixels from the raw RGB image, with the windows sampled at different image scales. The data is propagated through multiple fully connected layers and the output of the classifier is a probability of a given pixel belonging to the class fruit/non-fruit. The multi-scale patch representation of each pixel provides scale invariance for classification and allows us to capture local variations at different scales such as edges between fruits and leaves and between the trees and the skyline, while keeping the input dimensions low.

A three layer ms-MLP architecture is illustrated in Figure \ref{fig:mlp}. Given an image $I_k$, we are interested in finding the label of each pixel at location $(i,j)$ in the image. The input to the network are contextual patches $I_{i,j,k}^{s}$ surrounding the pixel at location $(i,j)$ in the $k^{th}$ image over scales $s\in\{1, \dots, S\}$. The image patches are initially forward propagated into the first hidden layer using non-linear sigmoid transformations. The activation outputs are then concatenated over the different scales:

\begin{equation}
H_1 = \bigcup_{s=1}^{S} \sigma(W_1^s I_{i,j,k}^{s} + b_1^s) 
\end{equation}

where, $\sigma(z) = 1/(1+e^{-z})$ is the sigmoid activation function. The weights for the individual scales $W^s$ are a set of linear filters/dictionaries transforming the input data into a more discriminative space. As done in \cite{Farabet2013}, input patches from each scale are treated independently. The concatenated first layer output $H_1$ is then propagated through subsequent densely connected layers:

\begin{figure}[h]
	\centering
		\includegraphics[width=1\textwidth]{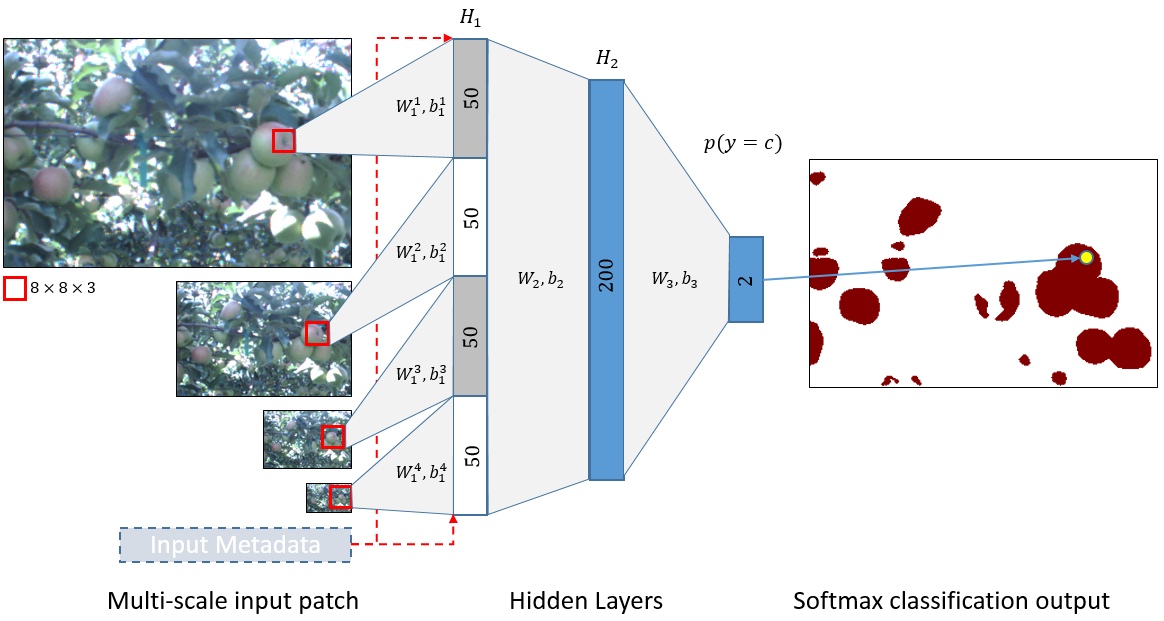}
		\caption{The multi-scale Multi-Layered Perceptron architecture with three layers. Centred at individual pixels, a contextual window is extracted over multiple image scales and the raw data is forward propagated through multiple hidden layers. The output softmax layer returns the probability of that pixel belonging to a given class. The optimal layer sizes are annotated on the individual layers above. Metadata are appended to the input layer, and their associated weights to the hidden layer are learnt during the training phase (red dashed line).}
	\label{fig:mlp}
\end{figure}

\begin{equation}
\label{eq:fclayer}
H_m = \sigma(W_m H_{m-1} + b_m) 
\end{equation}
for $m = \{ 2, ..., M \}$. The final layer is propagated through a softmax regression layer to obtain a class probability for the pixel $(i,j)$ belonging to the fruit or non-fruit class. 

\begin{equation}
\label{eq:softmax}
P(y_{i,j}=c|H_{M-1}) = \frac{e^{W_M^c H_{M-1}}}{\sum_{l=1}^{2} e^{W_M^l H_{M-1}}}
\end{equation}

 where $W_M^l$ are the final layers weights corresponding to class $l$. All parameters $(\textbf{W,b})$ of the network are learned in an end-to-end supervised way, by minimising a cross-entropy loss function. An $L_2$ regularisation penalty term is used to minimise over fitting. Optimisation is done with Stochastic Gradient Descent (SGD) using momentum and a linearly decaying learning rate.  

It has been shown in literature (and through our own experimentations) that unsupervised pre-training boosts classification performance for fully connected networks as they learn generalised features, which are useful for initialising the supervised training \citep{Erhan:2010}. Each set of first layer filters $W_1^s\ \forall s\in\{1, \dots, S\}$ are therefore pre-learnt using a held out dataset using a De-noising Auto encoder (DAE) with a sparsity penalty (see \cite{Hung2013} for details). The learnt weights consist of a combination of edge and colour filters. The deeper layers are initialised by using the sparse initialisation scheme proposed in \cite{Martens2010}.

\subsection{Convolution Neural Networks}
CNNs are feed forward neural networks, which concatenate several types of forward propagating layers, with convolutional layers playing a key role. Like the MLP, the network computes the probability of a pixel being a fruit or non-fruit, using as input the image intensities of a square window centred on the pixel itself. However, instead of using small multi-scale patches, CNNs can take as input larger/high resolution patches covering the same contextual region. This is due to the ability of the CNNs to share smaller scale filters in each layer, minimising the number of model parameters. 

\begin{figure}[!htb]
	\centering
		\includegraphics[width=1\textwidth]{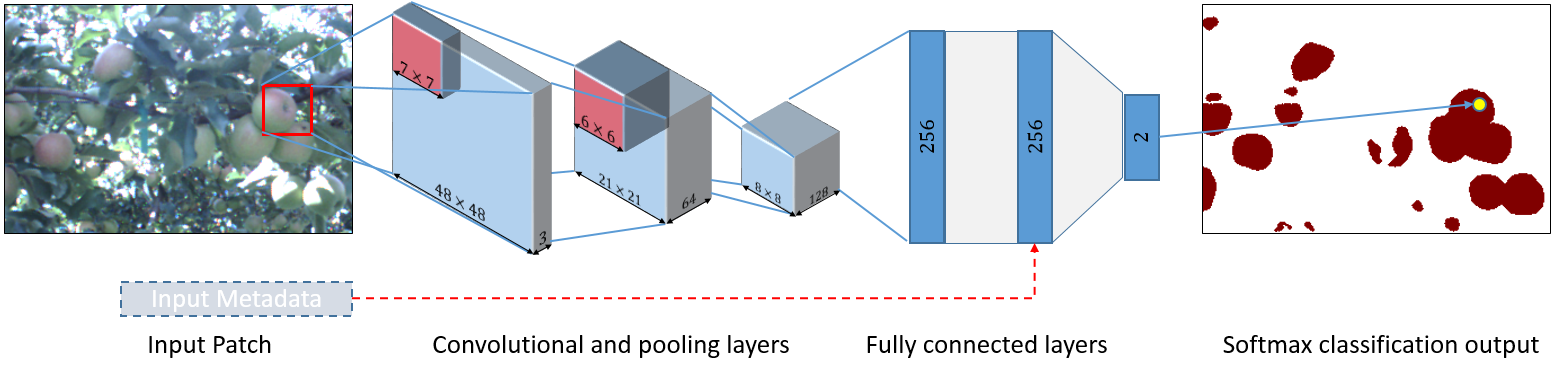}
		\caption{The patch based Convolutional Neural Network architecture with 2 convolution\texttt{+}pooling layers and 2 fully connected layers. Each pixel is defined by a large contextual region around it ($48\times48$ color pixels) and is propagated through the CNN. The output softmax layer returns the probability of that pixel belonging to a given class. The figure shows the optimal configuration for this dataset. Metadata are appended to the fully connected layer (red dashed line).}
	\label{fig:cnn}
\end{figure}

A typical CNN consists of a succession of convolutional, max-pooling and fully connected layers as shown in Figure \ref{fig:cnn}. Each convolutional layer performs a 2D convolution of its input maps with a square filter. This is followed by a non-linear activation function using the Rectified Linear Unit (ReLU) and a max-pooling sub-sampling layer as used in \cite{Krizhevsky2012}. The output of each block is then:
\begin{equation}
H_m = pool(ReLU(W_m H_{m-1} + b_m))
\end{equation}
where $H_0 = I_{i,j,k}$. The output from each layer is a feature map with successive layers covering increasingly large receptive fields, while encoding more complex variations in the input data. The convolution and pooling layers are followed by fully connected layers as per Equation (\ref{eq:fclayer}), but using the ReLU activation function instead of the sigmoid. Using a softmax activation function for the last layer (Equation (\ref{eq:softmax})) we obtain the class probability of the pixel/image patch. The network is trained in a similar fashion to the ms-MLP, however, unsupervised pre-training is not required with CNNs. Instead, dropout, an additional regularisation penalty on the fully connected layers is introduced during training. 

\subsection{Adding Metadata}
Metadata corresponding to individual pixels can be incorporated to image segmentation architectures by appending the information to one of the fully connected layers. We can define each set of metadata for a given input instance $I_{i,j,k}$, by $d_{i,j,k}^n\ \forall n \in N$, where $N$ is the set of different meta-parameters, e.g. sun position, tree type etc. The different metadata can then be concatenated together. 

\begin{equation}
D_{i,j,k} = \bigcup_{n=1}^N d_{i,j,k}^n
\label{eq:metadata-combine}
\end{equation}

For the ms-MLP, the propagation to the first hidden layer is then given by:

\begin{equation}
H_1 = \bigcup_{s=1}^{S} \sigma(W_1^s I_{i,j,k}^s + U^s D_{i,j,k}+ b_1^s) 
\label{eq:metadata-mlp}
\end{equation}

where, $U^s$ are the set of weights learnt for each scale for the scale independent metadata input $D$. The metadata have an additive effect on the biases $b_1^s$, where each metadata component shifts the responses from individual filters learned over the same layer. The metadata can equally be appended to deeper layers of the network, or first propagated through a disconnected hidden layer, however, for the ms-MLP our experiments found best performance when raw metadata were merged with the input layer. 

On the other hand, for the CNN architecture the metadata are added as inputs to a single fully connected layer (see Figure \ref{fig:cnn}) as: 
\begin{equation}
H_{m_{fc}} = ReLU(W_{m_{fc}} H_{m_{fc}-1} + UD + b_{m_{fc}})
\end{equation}

Both networks can be trained using the same back-propagation algorithm as before. The computational expense is linear to the number of nodes in a fully connected layer. Since the dimension of the metadata are significantly smaller than the size of the layers, the additional computational expense of this configuration is negligible. 

\subsection{Scene Inference}
\label{sec:method-inference}

During prediction, given a new image patch $I_{i,j,k}$, we can use feed forward propagation to predict the probability $p(y_{i,j}=fruit)$ for the pixel $(i,j)$ to belong to the fruit class. In practice, for both the ms-MLP and the CNN architecture, this can be performed densely via sliding windows to obtain a segmented image, but this approach is highly inefficient. Instead, a more computationally efficient approach is to re-design the learnt models as fully convolutional operations enabling fast dense prediction over arbitrary sized inputs. 

For the ms-MLP we transform the first layer weights $W_q^s$ into patch-wise kernels, which are convolved over the test images sampled at different scales $s\in \{1, \dots, S\}$. The filter responses from the smaller scales are up-sampled using linear interpolation to the same scale as the original image. The different scales are then concatenated and forward propagated through the subsequent fully connected layers. Zero-padding is applied to the input image to manage border effects from the convolution operations.  

For the CNN architecture, the convolution and pooling operations are easily scalable to larger images than the training input patches. The fully connected layers can also be seen as $1\times 1$ convolutions in a spatial setting, making the entire CNN simply a sequence of convolutions, activations and pooling operations. 

However, the output dimensions of CNNs are typically reduced due to the pooling layers, which sub-sample the data in order to keep the filters small and the computational requirements reasonable. As a result, the segmentation output from a network reduces the size of the input by a factor equal to the pixel stride of the pooling operations. For example, if the network includes two $2\times2$ pooling layers with a stride of $2$ pixels each, only $1$ in every $4$ pixels of the input image will be labelled. Some work with CNNs \citep{Farabet2013,Ning2005} upscale the label output by the sub-sampling factor to get to the input image size. 

To obtain a full resolution segmentation output without interpolation, input shifting and output interlacing is  used. If the network down-samples by a factor of $f$, we shift the input image (after zero padding) in the x and y directions by $\Delta_x, \Delta_y \in \{0, \dots, f-1 \}$. This results in $f^2$ inputs, each of which are passed through the CNN architecture producing down-sampled segmentation outputs. These are then interlaced such that the fine resolution predictions correspond to the pixels at the centres of their receptive fields \citep{Long2015,PinheiroC13,sermanet2013overfeat}.

\section{Experimental Setup}
\label{sec:setup}

\begin{figure}[h]
	\centering
	\begin{subfigure}[b]{0.32\textwidth}
		\includegraphics[width=\textwidth]{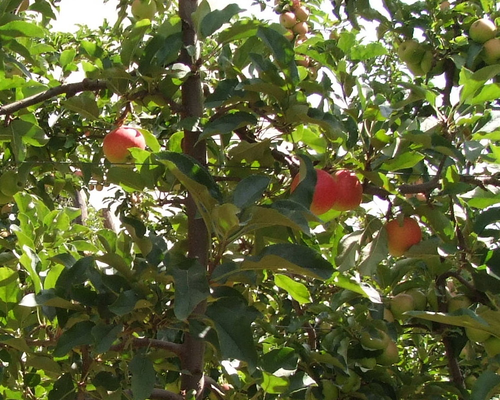}
		\caption{Kanzi (red apples)}
		\label{fig:apple-variety-kanzi}
	\end{subfigure}
	\begin{subfigure}[b]{0.32\textwidth}
		\includegraphics[width=\textwidth]{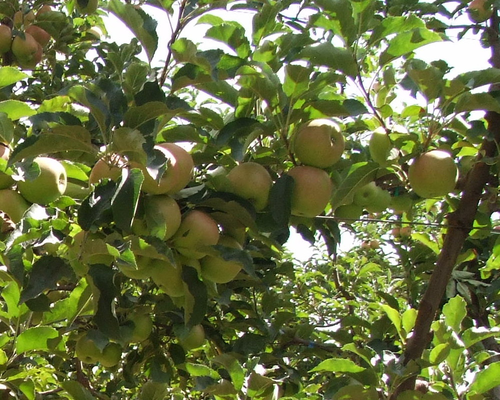}
		\caption{Pink Lady (pink-green apples)}
		\label{fig:apple-variety-plr}
	\end{subfigure}
	\begin{subfigure}[b]{0.32\textwidth}
		\includegraphics[width=\textwidth]{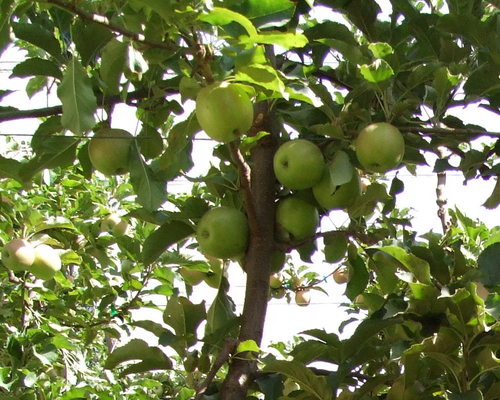}
		\caption{Pink Lady (green apples)}
		\label{fig:apple-variety-plg}
	\end{subfigure}
	\caption{Apple varieties found within the scanned block at the apple orchard. Their appearance ranges from bright red (Kanzi) to pink-green and leafy green (variants of Pink Lady). Sample images taken using a hand held DSLR camera.}
	\label{fig:apple-varieties}
\end{figure}

The orchard image segmentation architectures were tested at an apple orchard in Victoria, Australia (Figure \ref{fig:shrimp-orchard}). Data were gathered over a $0.5\ ha$ block, which hosts a modern G\"{u}ttingen V-trellis structure, where the crops are planted over pairs of trellis faces arranged to form a `V' shape. Built with support poles and wires along the rows, these structures provide better support of the tree limbs. They also allow for more sunlight for the fruits and easier harvesting. The trees hosted different apple varieties including Kanzi and Pink Lady. The fruit colour ranged from bright red to a mixture of pink-green and completely green as shown in Figure \ref{fig:apple-varieties}. The experiments were conducted one week before harvest (March 2013) where the apple diameter ranged from $70$ to $76\ mm$.

\subsection{Data Capturing}
The testing platform \textit{Shrimp} is a perception research ground vehicle, built at the Australian Centre for Field Robotics at The University of Sydney (Figure \ref{fig:shrimp-orchard}). Amongst an array of sensor types, it is equipped with a Point Grey Ladybug3 Spherical Digital Video Camera containing six $2$MP cameras oriented to capture a complete $360^{\circ}$ panoramic view.

To collect the dataset, the vehicle was teleoperated between $15$ orchard rows at $1.5\mbox{ to }2\ ms^{-1}$. The operator walked behind the vehicle, manually guiding it along the centreline of the rows, which were planted $4m$ apart. Images of size $1232\times1616$ were captured at $5$ Hz through the camera facing the trellis structure. The field of view of the camera was critical in its selection as it was able to capture the $4m$ tall trees from a distance of $2m$ as illustrated in Figure \ref{fig:shrimp-orchard}. The data were collected under natural illumination conditions, and the particular times of day, sun-angles and weather conditions were not specifically selected to optimise illumination. Finally, an on-board Novatel SPAN Global Positioning Inertial Navigation System (GPS/INS) provided estimates of the vehicle position and pose, which was used to localise each image. The vehicle trajectory is illustrated on the left in Figure \ref{fig:flow-chart}. 

We have also experimented with autonomous centre-line following using lidar for subsequent datasets, however, manual control yields data more similar to what would be expected from a manually driven tractor, which is one likely adoption strategy for the technology. Fully autonomous row following is likely to yield more consistent imagery, which improve image classification performance.

\subsection{Image Dataset}

\begin{figure}[h]
	\centering
		\includegraphics[width=\textwidth]{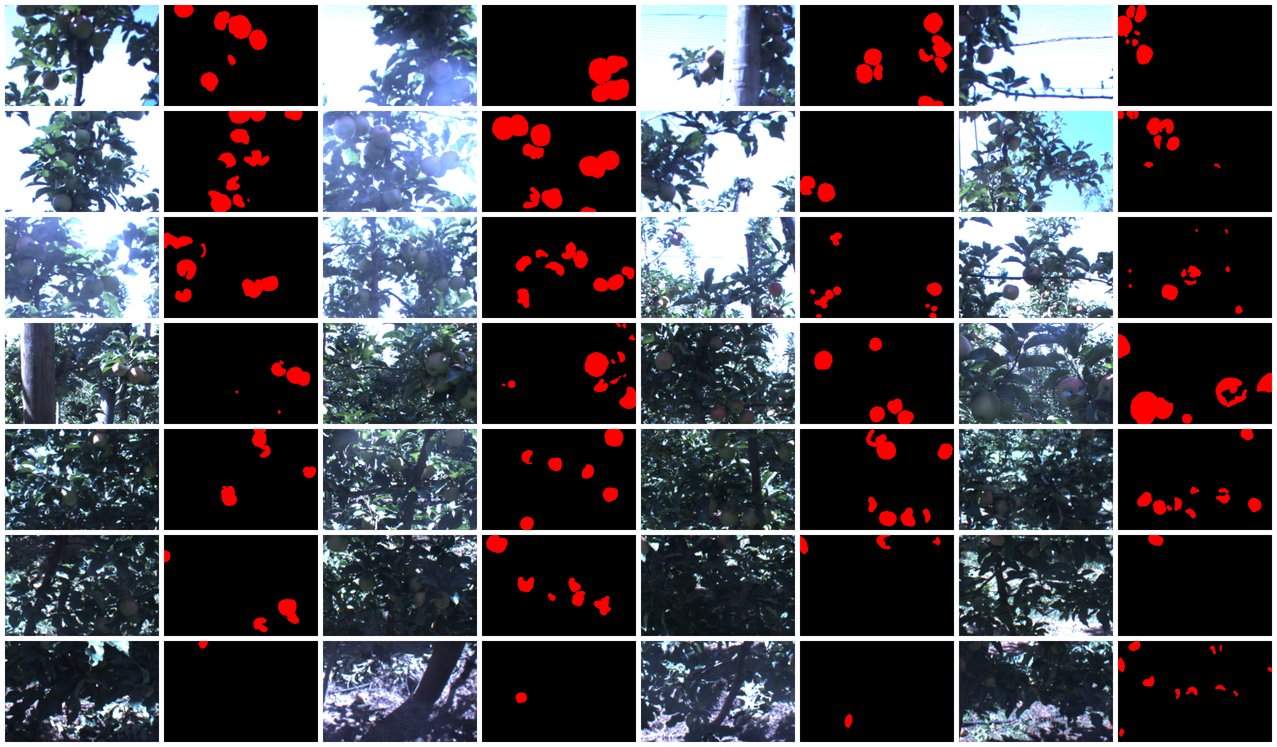}
		\caption{Sample sub-images randomly extracted from the orchard dataset for training. Images are vertically stacked in the figure according to their true height in the original data.}
	\label{fig:sub-tree-images-height}
\end{figure}

In total the image dataset consists of over $8000$ images of $1232\times1616$ pixels each. As it would be impractical to manually label this much data, a subset was collected by random sub-sampling. Each image was divided into $32$ sub-images with $308\times202$ pixels and a fixed number of sub-images were randomly sampled from each row to maximise the diversity of the data. In total $1100$ images were collected and manually annotated with binary pixel-level labels for the fruit and non-fruit class.  Sub-sectioning the image into smaller sections makes the manual labelling process easier and results in greater spatial variance within the dataset. Examples of the reduced dataset and their associated pixel-wise labels are shown in Figure \ref{fig:sub-tree-images-height}. Apples in the images varied in size from $25$ to $50$ pixels in diameter. This variation is attributed to the distance to the imaging platform and viewing angle.

\subsection{Segmentation Architecture}

\begin{figure}[h]
	\centering
		\includegraphics[width=\textwidth]{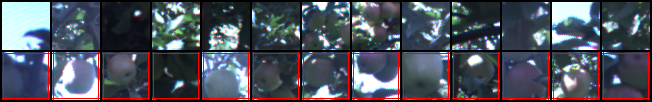}
		\caption{Image patches ($48\times48$) sampled from the labelled data to be used as input to the CNN architecture. Patches in the first and second rows are labelled 'not-apple' and 'apple' respectively.}
	\label{fig:image-patches}
\end{figure}

For training the ms-MLP, a similar architecture to the one presented in \cite{Hung2015} was used. Individual instances/patches of size $[8\times8\times3]$ were randomly sampled over image scales: $[1, 1/2, 1/4, 1/8]$. A separate natural scene dataset \citep{Brown2011} was used to initialise the first layer filters with a DAE. ZCA whitening was used for pre-processing. For the CNN, lacking equivalent implementation in agrovision, we build a network around previous work done for patch wise classification over electron microscopy images \citep{Ciresan2012} and urban scenes \cite{Brust15:CPN}. 
Pixel centred, single scale patches are extracted from the labelled dataset, while ensuring that they are large enough to identify/contain the fruits. Examples of $48\times48$ patches are shown in Figure \ref{fig:image-patches} with the corresponding labels denoting the class of the centre pixel. 

For all architectures, balanced sampling was done over the two classes due to the large class imbalance in the dataset. Hyperparameters such as the learning rate, learning rate decay factor, initial momentum, $L_2$ penalty, number of epochs, architecture width and depth were optimised over a held out validation set. The architectures were developed in Python using the open-source deep learning library, Pylearn2 \citep{pylearn2_arxiv_2013}. 

The metadata used for this dataset included a combination of pixel positions, orchard row numbers and the sun's position relative to the vehicle body frame. The sun's azimuth and elevation were calculated using the time of day and vehicle's pose and geographical position. The sun's elevation remained fairly constant over this dataset, therefore only the azimuth was considered. These properties were chosen as they were qualitatively observed to correlate with some of the intra-class variation in the data. For example, Figure \ref{fig:sub-tree-images-height} illustrates the variation in appearance as a function of the vertical position in the captured image, with the image regions higher up on the tree appearing brighter and hosting apples viewed from a different pose. 

The combined metadata information (Equation (\ref{eq:metadata-combine})) was incorporated within the ms-MLP and the CNN as shown by the dashed lines in Figures  \ref{fig:mlp} and \ref{fig:cnn} respectively. Each meta-parameter can be formatted as either a single continuous unit or multiple discrete units. For example, the height of a pixel on the image (in the range $[0-1616]$) can be normalised and represented by a single number in the range $[0-1]$ or represented as a one-hot encoded vector. In the latter, image height is discretised into a fixed number of bins with a single active unit denoting the pixel height (as done in \cite{Rao2014}). However, a single unit representation restricts the number of parameters available to represent the metadata and prevents us from learning a different set of biases of individual filters responses over the variations within each meta-parameter. A one-hot encoding was therefore used, where a few channel sizes were tested and ultimately $8$ channels were used for each continuous input data: pixel $i$, $j$ positions ($p_i, p_j$) and the sun azimuth angle ($s_\psi$). The row numbers ($r_n$) could be encoded directly as one-hot vectors. To test how the learning algorithm would cope with irrelevant metadata information, training of the ms-MLP architecture was also conducted with the inclusion of uniformly distributed random noise as information. No further hyperparameter optimisation was performed following the inclusion of the metadata.

\section{Segmentation Results}
\label{sec:seg-results}

To evaluate the image segmentation performance, the labelled dataset ($1100$ images) was randomly divided into an $80-10-10$ split of training, validation and testing images. 
Single scale and multi-scale patches were sampled (balanced between the classes) from the set of training images to train the CNN and ms-MLP architecture \footnote{Due to spatial correlation in appearance, dense sampling of training instances is not necessary.}. In this section we evaluate the performance of the different architectures using the fruit F1-score evaluated over instances randomly sampled from the test image set. The class threshold for the optimal F1-score was evaluated over the instances sampled from the validation set. In all tests, mean and one-standard deviation classification results were obtained over 10 iterations, while shuffling the $80-10-10$ split. The last part of this section utilised the patch based models for whole image inference, presenting qualitative segmentation results over images from the test set. 

\subsection{Multi-scale Multi-Layered Perceptron}

For baseline comparison, the original architecture from \citep{Hung2015}, a 2-layer MLP with $200$ hidden units was trained with $200,000$ training instances, which we denote as ms-MLP-2. Different sources of metadata were then added to this configuration, and the MLP retrained each time. Training each MLP network till convergence over the validation set ($\sim30$ epochs) took roughly $5$ minutes on a GPU enabled desktop. The classification results are shown in Table \ref{tab:mlp-pixel-results}. 

\begin{table}[h]
\caption{Pixel-wise fruit classification results with the original ms-MLP architecture as presented in \cite{Hung2015}. Combinations of metadata are added to the network, and the corresponding classification results listed as the absolute F1-score and as difference from the default method.}
\label{tab:mlp-pixel-results}
\begin{center}
\begin{threeparttable}{
\begin{tabular}{|c|c|c|}
\hline
NN Architecture & Metadata & Pixel F1-score \\
\hline
\multirow{8}{*}{ms-MLP-2} & None & $0.683 \pm 0.015$ \\
\cline{2-3}
 & Noise & $0.683\ (+ 0.000 \pm 0.002)$ \\ 
\cline{2-3}
 & $p_i$ & $0.715\ (+ 0.032 \pm 0.005)$ \\
\cline{2-3}
 & $p_j$ & $0.683\ (+ 0.000 \pm 0.002)$ \\
\cline{2-3}
 & $r_n$ & $0.694\ (+ 0.011 \pm 0.004)$ \\
\cline{2-3}
 & $s_\psi$ & $0.684\ (+ 0.001 \pm 0.003)$ \\
\cline{2-3}
 & $p_i$, $r_n$ & $0.721\ (+ 0.038 \pm 0.005)$ \\
\cline{2-3}
 & $p_i$, $p_j$, $r_n$, $s_\psi$ & $0.725\ (+ 0.042 \pm 0.005)$ \\
\hline
\end{tabular}
\begin{tablenotes}
\scriptsize
\item Pixel position ($p_i, p_j$), row number ($r_n$), sun azimuth ($s_{\psi}$)
\end{tablenotes}}
\end{threeparttable}
\end{center}
\end{table}

There is a clear improvement in classification results with the inclusion of all of the metadata increasing the F1-score the most by $6.2\%$ ($F1: 0.683 \rightarrow 0.725$). On it's own, $p_i$ (the height within the original image) was found to be the most important individual metadata parameter, which was also empirically the most related to appearance variations in the data. Additionally, if some metadata were hypothesised to have no direct relationship with the extrinsic variations in data (such as random noise or $p_j$), the classification results were no worse. The lack of information from the sun position $s_{\psi}$, is possibly due to the fact that in this dataset, the sun azimuth angle was either $+90^{\circ}$ or $-90^{\circ}$, and therefore did not cause much variation on its own to captured images. Interestingly, the learnt weights $U$ from Equation (\ref{eq:metadata-mlp}) corresponding to these metadata had a much weaker response, meaning that during the learning phase, the algorithm discovers which inputs are not relevant to the classification task. Additionally, this analysis guides us towards the most important factors of variation, which in turn might be physically controllable, to improve subsequent data collection. 

\subsubsection{Optimal ms-MLP Architecture}

We extend from \cite{Hung2015} by searching over different combinations of depth and width for the MLP architecture. A grid search was performed over networks with a depth of 2-5 layers and with varying width of 200 to 1000 units per layer. The optimal fruit classification performance was obtained with a 3-layer MLP with each hidden layer containing 200 hidden units. We denote this network as ms-MLP-3* and it is illustrated in Figure \ref{fig:mlp}. As before, metadata were then added within this network at the input layer. During the optimisation phase we experimented with adding the metadata to deeper layers instead of the input layer and propagating it through independent hidden layers before merging with the image data as done in \cite{Rao2014}. However, the best results were obtained when the raw metadata were added alongside the input multi-scale image data. The segmentation results are reported in Table \ref{tab:mlp-pixel-results-optimised}. Going from the ms-MLP-2 to the optimised ms-MLP-3* network, the F1-score increases from $0.683$ to $0.728$. With the inclusion of metadata (using all meta-parameters) there is a statistically significant improvement in the F1-score by $+0.023$ to $0.751$. The magnitude of the improvement is smaller, possibly due to the higher model complexity automatically capturing some of the appearance variations and class distributions in the data. 

\begin{table}[h]
\caption{Pixel-wise fruit classification using different ms-MLP architectures. The networks used are ms-MLP-2 from \cite{Hung2015} and ms-MLP-3*, optimised over width and depth. Metadata results are listed for both networks as the absolute F1-score and as change from the configuration not using metadata.}
\label{tab:mlp-pixel-results-optimised}
\begin{center}
\begin{threeparttable}{
\begin{tabular}{|c|c|c|}
\hline
NN Architecture & Metadata & Pixel F1-score \\
\hline
\multirow{2}{*}{ms-MLP-2} & None & $0.683 \pm 0.015$ \\
\cline{2-3}
 & $p_i$, $p_j$, $r_n$, $s_\psi$ & $0.725\ (+ 0.042 \pm 0.005)$ \\
\hline 
\multirow{2}{*}{ms-MLP-3*} & None & $0.728 \pm 0.016$ \\
\cline{2-3}
 & $p_i$, $p_j$, $r_n$, $s_\psi$ & $0.751\ (+ 0.023 \pm 0.008)$ \\
\hline
\end{tabular}
\begin{tablenotes}
\scriptsize
\item Pixel position ($p_i, p_j$), row number ($r_n$), sun azimuth ($s_{\psi}$)
\end{tablenotes}}
\end{threeparttable}
\end{center}
\end{table}

\subsubsection{Varying Training Size}

\begin{figure}[h]
	\centering
	\includegraphics[width=0.7\textwidth]{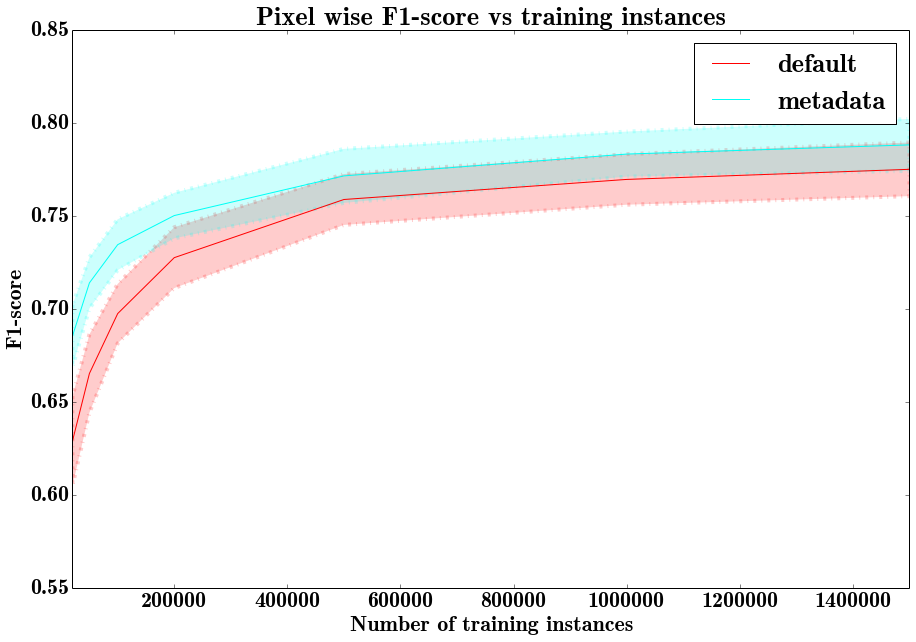}
		\caption{Classification results against number of training instances with and without metadata over the two segmentation architectures. The shaded regions illustrate one standard deviation in the results.}
	\label{fig:num-instances}	
\end{figure}

As mentioned in Section \ref{sec:intro}, a classifier is more likely to be invariant to intra-class appearance variations given a high model complexity and enough training data. In the previous section we showed this to be true when the model complexity was increased. The training size for the patch based segmentation architectures is the number of pixels (neighbourhood patches) extracted from the set of labelled images for training. A larger number of training instances would typically result in better classification performance but results in a linear increase in algorithm run-time during training. Using the ms-MLP-3* configuration, we evaluate the effects of training size on the classification results, with and without the inclusion of metadata.  The classification F1-score (shown in Figure \ref{fig:num-instances}) increases as we sample more instances from the training set, reaching convergence around $500,000$ training instances. Theoretically, a total of $\sim 56\times10^6$ patches can be sampled from the $900$ training images, however, there can be significant overlap in information due to the contextual area covered by single training instances. The addition of the metadata to the architecture results in improved classification performance in all cases, having a greater advantage when restricted with the number of training instances. Additionally, for a fixed F1-score, the inclusion of metadata allows the same result with only a small fraction of the training instances (e.g. the same F1-score can be achieved with the metadata model while using just $200,000$ training instances, compared to $500,000$ instances when no metadata are used). As the classification performance converges, the performance improvement due to metadata converges to approximately $+0.013 \pm 0.007$.

\subsection{Convolutional Neural Network}
For the optimal CNN configuration, we searched over different depths, widths and input sizes while monitoring the F1-score over the held out validation set. We tested a range of input sizes, covering a $\{32\times 32\}, \{48\times 48\}, \{64\times 64\}$ region around a pixel. For the convolutional layers, we tested a depth of 2-3 layers, each containing a convolution, activation and pooling operation. These were followed by two fully connected layers as typically done in computer vision literature. The optimal architecture was found by doing a coarse grid search and is denoted in this paper as CNN* (illustrated in Figure \ref{fig:cnn}). The first convolutional layer filters the $48\times 48\times 3$ input image patch with $64$ kernels of size $7\times 7\times 3$ and is followed by a ReLU activation and a max-pooling operation over a $2\times 2$ patch with a stride of $2$ pixels. The second convolutional layer filters the sub-sampled $21\times 21\times 64$ input with $128$ kernels of size $6\times 6\times 64$ followed by a similar activation and pooling layer. The resulting $8\times8\times128$ output is propagated through two fully connected layer with $256$ nodes each and finally a softmax layer.

Training the CNN architecture till convergence over the validation set ($\sim50$ epochs) took roughly $3$ hours while using $200,000$ training instances. The CNN patch classification results are shown in Table \ref{tab:cnn-pixel-results}. Without metadata, the F1-score has increased from $0.728$ (ms-MLP-3*) to $0.791$ with the CNN* architecture. 

\begin{table}[h]
\caption{Pixel-wise fruit classification results with the CNN* architecture using the default method and with the addition of Metadata. Metadata results listed as the difference in F1-score from the default method.}
\label{tab:cnn-pixel-results}
\begin{center}
\begin{threeparttable}{
\begin{tabular}{|c|c|c|}
\hline
NN Architecture & Metadata & Pixel F1-score \\
\hline
\multirow{2}{*}{CNN*} & None & $0.791 \pm 0.011$ \\
\cline{2-3}
 & $p_i$, $p_j$, $r_n$, $s_\psi$ & $0.797\ (+ 0.006 \pm 0.008)$ \\ 
\hline 
\end{tabular}
\begin{tablenotes}
\scriptsize
\item Pixel position ($p_i, p_j$), row number ($r_n$), sun azimuth ($s_{\psi}$)
\end{tablenotes}}
\end{threeparttable}
\end{center}
\end{table}

As before, we add the metadata to the classification architecture, which was optimally placed alongside the first fully connected layer. In contrast to what was observed with the ms-MLP architectures, the increase in F1-score with the inclusion of the metadata to CNN* is minimal at $+0.006$ (Table \ref{tab:cnn-pixel-results}). Additionally, in practice we observed that in some configurations (e.g. if the metadata were added to the second fully connected layer), the performance was slightly worse. 

\subsection{Whole Image Segmentation}
To acquire the input for consequent fruit detection and yield estimation, and to qualitatively analyse the segmentation results, the trained ms-MLP and CNN architectures were utilised to segment whole images using the techniques discussed in Section \ref{sec:method-inference}. The image inference was conducted using the ms-MLP-3* architecture with and without metadata and with the CNN* architecture without metadata. Inclusion of metadata with the CNN architecture was not tested here due to the minimal classification gain it provided. 

\begin{figure}[h]
	\centering
		\includegraphics[width=\textwidth]{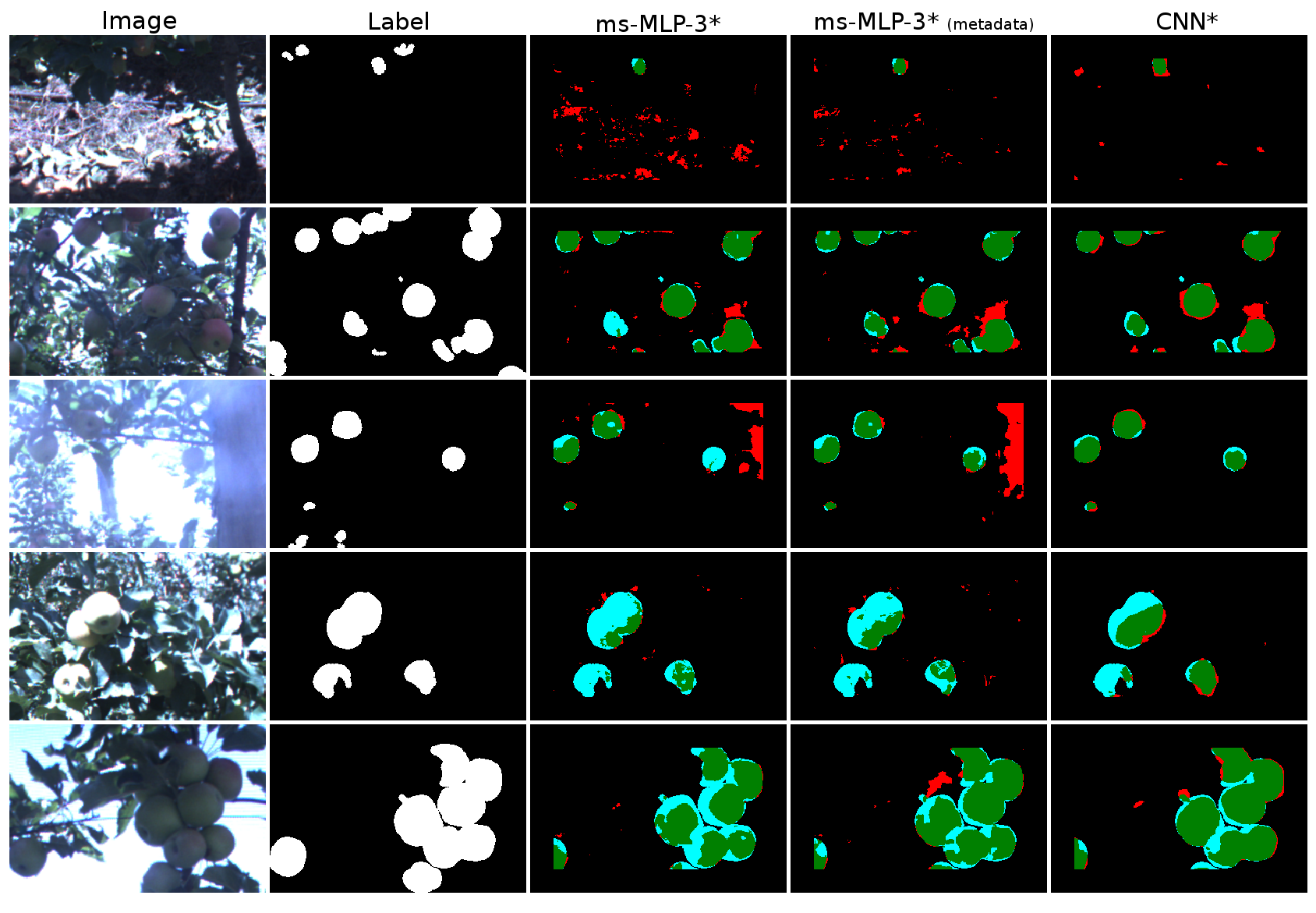}
		\caption{Image segmentation results. Five sample image sections from the test image set (first column) and their ground truth labels (second column). The third and fourth columns illustrate the segmentation output from the ms-MLP-3* architecture without and with metadata. The fifth column shows the segmentation output from the CNN* architecture. The image borders are masked out during performance evaluation. The output segmentation is illustrated as, green: true positives, red: false positives, black: true negatives and cyan: false negative classifications.}
	\label{fig:clf-results}
\end{figure}

Whole image segmentation using the ms-MLP architecture on the $308\times202$ images from the test image set took $0.60$ seconds/image. The largest computational overhead is during the up-scaling stage of the smaller scale first layer filters. The resultant binary mask for a few images is shown in columns 3 and 4 in Figure \ref{fig:clf-results}, with and without metadata. The metadata configuration is picking a greater region of apples and even some new apples that the no-metadata configuration does not detect (e.g. centre right apple in the third row and centre left apple in the fourth row). Additionally, over certain regions it has a much lower false positive rate as seen in the top row example. However, there are some instances where adding metadata lowers precision as seen by the false detection of the trunk in the third row. 

The CNN based image segmentation is performed using the shift and stitch method which took $0.24$ seconds/image on the $308\times 202$ images. The resultant binary mask is shown in the last column in Figure \ref{fig:clf-results}. As expected from the performance metrics, we see the CNN* network outperforming the others in both segmentation precision and recall. For example, even under the challenging illumination example, the third row trunk is not miss-classified. The improved performance could be attributed to the higher resolution input and the greater number of available parameters to capture the data distribution. Finally, we observe a smoother results with the CNN, which is due to the pooling layers enforcing some translation invariance in the data. 

\section{Fruit Detection and Yield Prediction}
\label{sec:detection}

The output from the image segmentation methods above are class probability maps or binary masks indicating the location of individual fruit pixels in the images. In this section we present methods to translate this to agronomically meaningful information, such as detection of individual fruits, fruit counts and maps of fruit yield. The segmentation architectures we focus on are the ms-MLP-3* (no metadata), the ms-MLP-3* with metadata and the CNN* (no metadata). The metadata configuration of the CNN network is not tested any further as it yielded minimal improvements in the segmentation results. Furthermore, we evaluate the fruit detection and yield estimation performance using different ground truth data, comparing the impact of different segmentation architectures towards higher level agricultural tasks. 

\subsection{Fruit Detection}

The standard approach for object detection in the computer vision literature is to first perform dense prediction using sliding windows over the image, score each window via a trained classifier (as done in the previous section) and then remove overlapping windows via non-maximum suppression. Training is done using image patches centred around objects of interest, therefore during prediction the probability maps peak near the centre of individual objects. However, for image segmentation, where patches are trained using the centre pixel class, the probability map plateaus around fruit region (i.e. no local maxima near the centre of the fruit). Therefore for fruit detection, we aim to split the binary fruit mask, obtained from the probability map, into separate regions denoting individual fruits. 

If individual fruits were spread apart over the segmented image and appeared without any occlusions, fruit detection would simply involve selecting the disjoint regions in the binary output. However, in orchards, fruits are often clustered together and appear occluded in the image data due to other fruits and foliage. A suitable fruit detection algorithm needs to be able to separate individual fruits in clusters and also connect disjoint regions of the same fruit, formed due to occlusions. For this, we implement two different detection techniques, the Watershed Segmentation (WS) algorithm \citep{Roerdink2000} and the Circular Hough Transform (CHT) algorithm \citep{Atherton1999}.

\begin{figure}[h!]
	\centering
		\includegraphics[width=\textwidth]{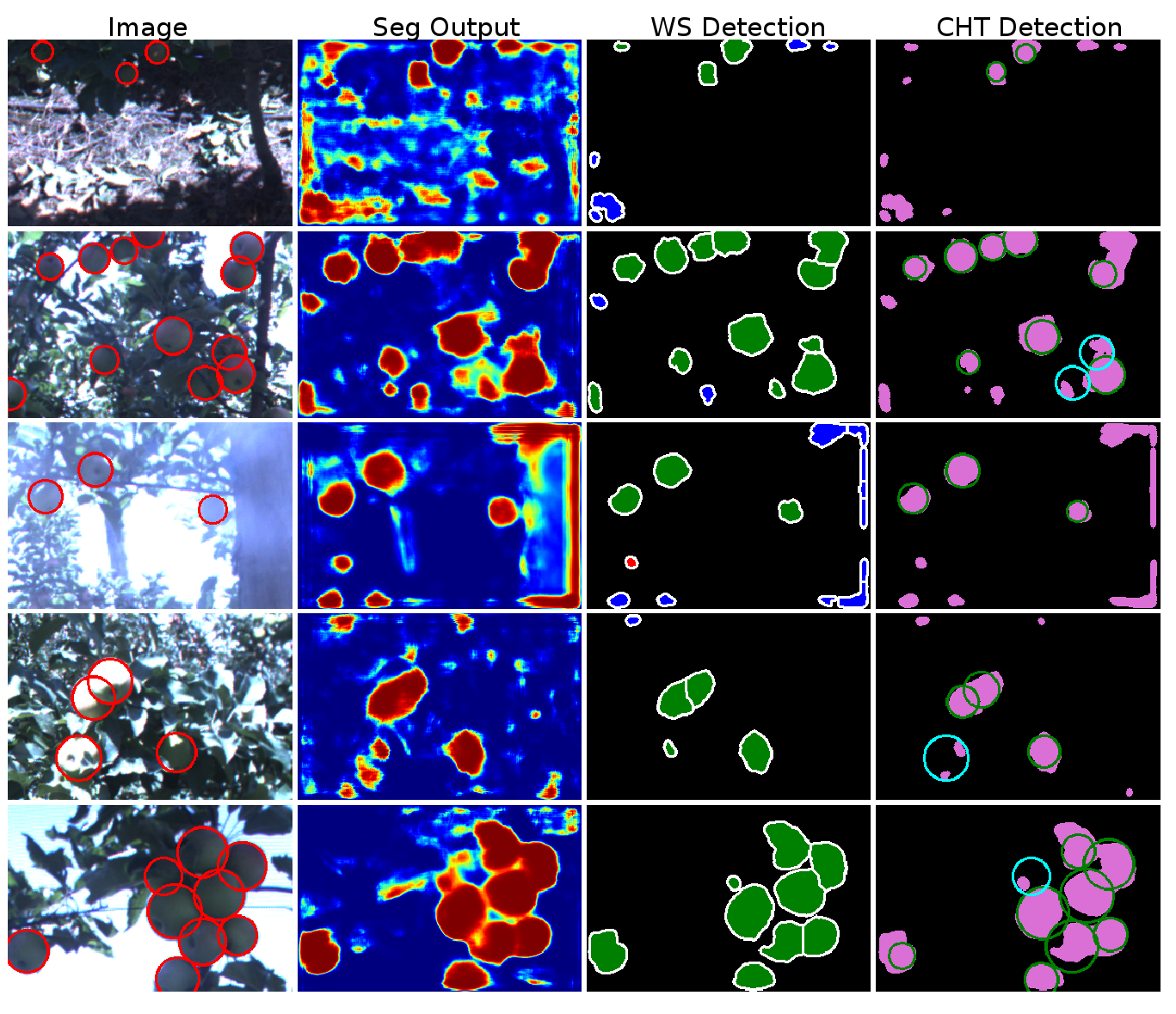}
		\caption{Apple detection results. The first column contains the orchard images with manually annotated fruit. The second column is the segmentation output from the CNN* algorithm. Individual fruit detection is done using the Watershed Segmentation and Circular Hough Transform algorithms, with the output shown in columns 3 and 4 respectively. Detections are shown as, green: true positive, red: false positive and cyan: false negative. Detections near the image boundaries (blue) are ignored during evaluation.}
	\label{fig:detection-results}
\end{figure}

The WS algorithm is used for separating connected objects in an image where each object is characterised by a local maximum and a contour leading to its boundary. For a segmented fruit image, this contour is evaluated by computing distances of individual fruit pixels to the nearest background pixel, resulting in a local maximum that is typically situated around the centre of each fruit. The detection output using the WS approach is shown in the third column on Figure \ref{fig:detection-results}. Individual fruits are discernible even if they appear in clusters. However, the algorithm cannot merge fragments of a single fruit that become separated due to foreground occlusion or misclassification. To overcome this, we also tested CHT on the binary segmentation output. The algorithm is reliant on the circular nature of the fruits and has the added advantage of detecting partially circular regions, enabling the potential to merge disjoint fruit regions into a single detection. Example of detections using the CHT are shown in the fourth column in Figure \ref{fig:detection-results}. For both detection operations, the segmentation output is pre-processed with morphological erosion and dilation to enforce local consistency. 

In order to evaluate fruit detection performance, we rely on ground truth available at the individual fruit level. This information can be extracted from the pixel-labelled data by doing fruit detection using the approaches mentioned above (as done in \cite{Bargoti2016ICRA}). However, this may induce a bias in the ground truth where clustered or occluded fruits are not accurately represented due to errors in the detection algorithms. Therefore in this paper, we annotate the same set of images with individual fruit labels, labelling them with a circular marker with variable radius, and storing the centre and radius of each fruit\footnote{The python based annotation toolbox is available at \url{https://github.com/sbargoti/pychetlabeller.git}}. Some of the annotations are shown in the left column in Figure \ref{fig:detection-results}. A greedy 1-nearest neighbour, one-to-one matching was performed between the ground truth and estimated detections, with a detection denoted as true positive if it was within the annotated fruit region. The remaining detections were classified as false positives. True fruit locations not associated with any detections were denoted as false negatives. The detection F1-scores using the WS and CHT algorithms on the segmentation outputs from three different architectures are shown in Table \ref{tab:detection-results}. For fruit counting, we evaluate the total number of fruit detections per image (true positives + false positives). The counting evaluation metric used is the coefficient of determination ($r^2$) between the detection counts and the annotated counts over the images in the test sets. The corresponding regression results for the different configurations are also shown in Table \ref{tab:detection-results}. All detection hyperparameters (e.g. minimum distances between fruits, fruit size) were evaluated over the validation image set\footnote{Some of the detection hyperparameters are related to the size of the fruit in the images, which will be extracted automatically in the future by evaluating the distances from the sensor to the fruit.}.

\begin{table}[h]
\caption{Individual fruit detection results (F1-score) using the Watershed Segmentation and Circular Hough Transform detection algorithms on the image segmentation output. The segmentation architectures are the ms-MLP-3* (without and with metadata) and the CNN* (without metadata).}
\label{tab:detection-results}
\begin{center}
\begin{tabular}{|c|c|c|c|c|}
\hline
\multirow{2}{*}{Segmentation Architecture} & \multicolumn{2}{|c|}{WS Detection} & \multicolumn{2}{|c|}{CHT Detection} \\
\cline{2-5}
& F1-score & r-squared & F1-score & r-squared \\
\hline
ms-MLP-3* (no metadata) & $0.853 \pm 0.010$ & $0.722 \pm 0.048$ & $0.818 \pm 0.019$ & $0.702 \pm 0.050$ \\ 
\hline 
ms-MLP-3* (with metadata) & $0.853 \pm 0.009$ & $0.742 \pm 0.054$ & $0.821 \pm 0.012$ & $0.704 \pm 0.041$ \\ 
\hline
CNN* & $0.858 \pm 0.010$ & $0.770 \pm 0.050$ & $0.857 \pm 0.014$ & $0.767 \pm 0.046$ \\ 
\hline 
\end{tabular}
\end{center}
\end{table}

The WS and CHT fruit detection algorithms were comparable and best performing with the CNN* architectures, with a detection F1-score of $0.858$. However, with the ms-MLP architectures, WS outperformed CHT detection. This could be due to a smoother segmentation result obtained with the CNN, where the binary fruit regions have rounder boundaries, which are more suitable for CHT detection. Interestingly, we did not observe statistically significant improvement in detection accuracy using the ms-MLP-3* with and without the metadata, despite the performance gains found at the pixel level. Additionally, with the WS algorithm, there is minimal increase in detection performance when using the CNN network over the ms-MLP networks. However, the regression scores vary between the outputs from the segmentation architectures, with the strongest r-squared fit of $0.770$ observed with CNN* and WS detection. 

\subsection{Yield estimation}

Accurate image segmentation and fruit detection can enable accurate yield estimation. However, even with perfect segmentation, followed by an appropriate fruit detection algorithm, we cannot directly observe the true number of apples due to occlusions. Instead, we make the assumption that on average there is a constant ratio of visible fruits to occluded fruits, allowing us to either map variations in the yield at the farm or perform yield estimation given some calibration data. At this apple orchard, the grower provided the calibration data by counting and weighing the post-harvest produce, separately per row\footnote{Harvest fruit weights and counts are normally measured per orchard block. Per-row counts are typically too labour intensive for a commercial orchard to routinely perform.}. The data covered $15$ rows, ranging from $3,000$ to $12,000$ apples per row.

\begin{figure}[!h]
	\centering
		\includegraphics[width=\textwidth]{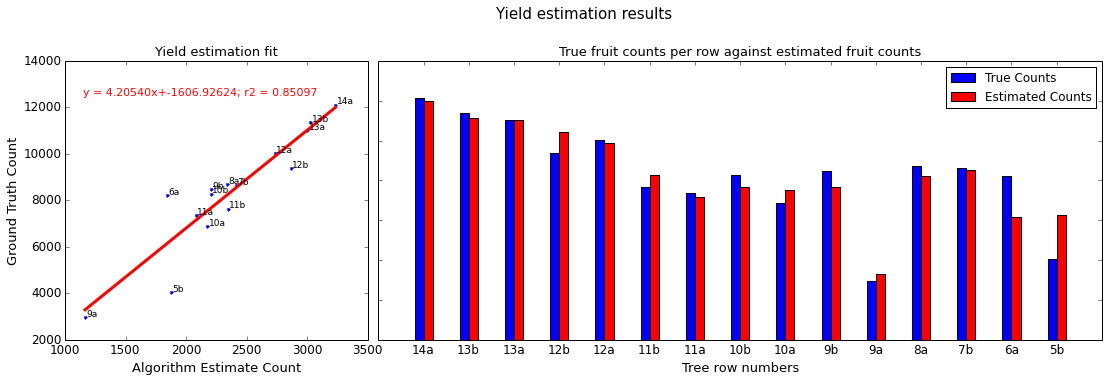}
		\caption{Yield estimation results. On the left is the predicted yield over $15$ rows fitted against the post-harvest ground truth per row. The estimated yield counts against the true counts are shown on the right.}
	\label{fig:yield-estimate}
\end{figure}

Yield estimation was performed using the original $1232\times1616$ images captured densely over the farm. The images were first downsampled to every $0.5\ m$ along the row to minimise image overlap (and hence double counting) in subsequent frames. Any remaining overlap was avoided by manually choosing a fixed region of interest along the image centre, within which the detected fruits were counted. Fruit counts were accumulated over the images per row to give an uncalibrated row count estimate. Similar to \cite{Hung2015}, yield estimation accuracy was evaluated using the r-squared correlation coefficient between the estimated row yield and the true counts as shown on the left in Figure \ref{fig:yield-estimate}. The evaluation was done over multiple image segmentation training iterations, using the three different segmentation architectures. The results are shown in Table \ref{tab:yield-results}.

\begin{table}[h]
\caption{Yield estimation results using different image segmentation architectures and fruit detection algorithms. The results are the r-squared fit between the true per-row counts and the estimation counts, and the average absolute error in the yield estimation over the different rows.}
\label{tab:yield-results}
\begin{center}
\begin{tabular}{|c|c|c|c|c|}
\hline
\multirow{2}{*}{Segmentation Architecture} & \multicolumn{2}{|c|}{WS Detection} & \multicolumn{2}{|c|}{CHT Detection} \\
\cline{2-5}
& r-squared & Error (\%) & r-squared & Error (\%) \\
\hline
ms-MLP-3* (no metadata) & $0.753 \pm 0.047$ & $13.3 \pm 1.25$ & $0.635 \pm 0.051$ & $17.53 \pm 1.14$ \\ 
\hline 
ms-MLP-3* (with metadata) & $0.771 \pm 0.036$ & $12.2 \pm 1.13$ & $0.705 \pm 0.072$ & $14.4 \pm 2.14$ \\ 
\hline
CNN* & $0.826 \pm 0.021$ & $10.84 \pm 0.62$ & $0.763 \pm 0.015$ & $14.15 \pm 0.32$ \\ 
\hline 
\end{tabular}
\end{center}
\end{table}

Across all segmentation architectures, the use of WS detection algorithm produced more accurate yield estimation results compared to the CHT algorithm. This is on par with what was observed with the detection and counting results from Table \ref{tab:detection-results}. Using the ms-MLP-3* configuration (without metadata) and WS detection, we achieved an r-squared value of $0.753$. This is greater than the baseline score of $0.635$ achieved using CHT detection, comparable to $0.656$ reported in \cite{Hung2015}\footnote{The previous publication mistakenly reported an r-squared value of $0.81$, which was in fact the r-value.}, evaluated over a single iteration using the ms-MLP-2 network and the CHT detection framework. As reported in \cite{Bargoti2016ICRA}, using the ms-MLP-3* architecture with metadata, the linear fit with WS detection increased to $0.771$. Finally, with the CNN* architecture, the r-squared increased to $0.826$, the best yet on this dataset. A linear fit between the algorithm row counts and the true row counts is shown on the left in Figure \ref{fig:yield-estimate}. 

The calibrated linear models relating the algorithm counts to true fruit counts can be used to estimate the yield in each row. The count estimates using the optimal configuration is shown on the right in Figure \ref{fig:yield-estimate}. Yield estimation error can be evaluated by accumulating the absolute estimation errors per row normalised against the total fruit count. The associated results are shown in Table \ref{tab:yield-results}. The use of metadata on the ms-MLP-3* architecture, improved this error from $13.3\%$ to $12.2\%$, whereas with the CNN* architecture, this improved to $10.84\%$.

\begin{figure}[!h]
	\centering
		\includegraphics[width=\textwidth]{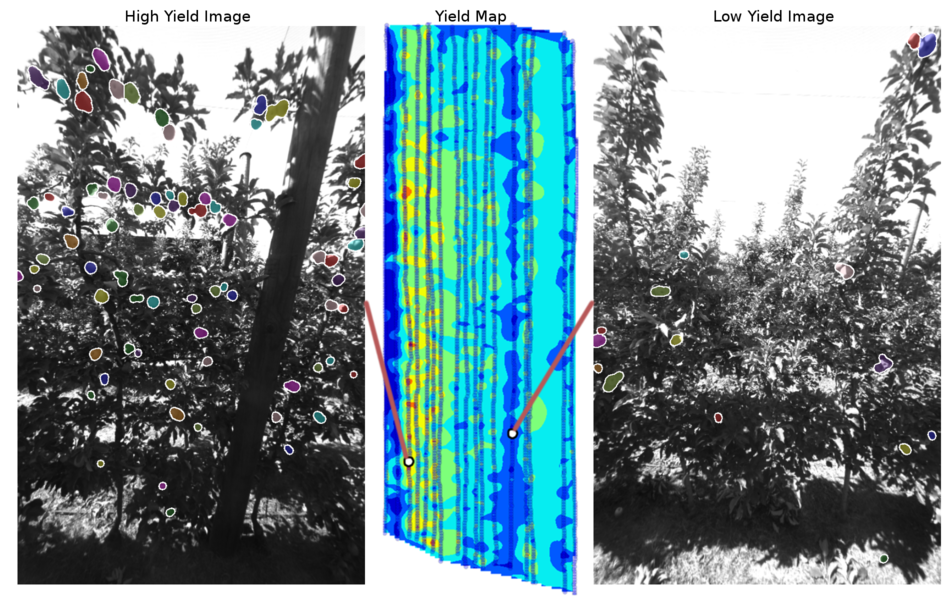}
		\caption{Apple yield map for the orchard block. Individual geo-referenced images are segmented using the CNN architecture and fruit detection is performed using CHT to obtain a fruit count per image. The yield map is computed through bilinear spatial interpolation of these counts. The insets show image samples with fruit detections in areas of high and low yield. }
	\label{fig:yield-map}
\end{figure}

Finally, a yield map can be produced by geo-referencing the counts per image given the vehicle position. The predicted counts can then be interpolated over a fixed grid to produce a yield map of the orchard block as shown in Figure \ref{fig:yield-map}. The yield map shows spatial variability in yield at the orchard block as illustrated by the figure insets, showing examples of areas with low and high yield count. 

\section{Discussion}
\label{sec:discussion}

Image segmentation accuracy for orchard image data increased when utilising more complex classification architectures. Operating under natural illumination conditions, the previously proposed ms-MLP network by \cite{Hung2015} resulted in a fruit segmentation F1-score of $0.683$. Optimising this architecture with a deeper network increased the score to $0.728$, which was further increased to $0.751$ with the inclusion of metadata, as previously shown in \cite{Bargoti2016ICRA}. With the state-of-the-art CNN approach, we obtain the highest F1-score of $0.791$, which is a modest improvement, yet importantly brings the state-of-the-art in agrovision for apple detection closer to state-of-the-art in computer vision generally. Qualitatively, we observed smoother segmentation results with the CNNs over the ms-MLP architectures as shown in Figure \ref{fig:clf-results}. The CNN architecture did not require any data pre-processing or layer-wise training prior to the supervised learning phase, however, training time was considerably slower compared to ms-MLP ($5$ min vs $3$ hrs). For prediction, the CNN model could be implemented as a fully convolutional operation, leading to $2-3$ times faster inference. With both architectures, it was important to conduct a thorough hyperparameter search over a held out validation dataset in order to obtain the optimal segmentation results.

Orchard metadata were incorporated into the different architectures to help model some of the appearance variations and/or class distributions observed in the data. The maximum improvement in pixel-classification F1-score with metadata was observed when operating with the sub-optimal ms-MLP-2 network (Table \ref{tab:mlp-pixel-results}) and/or with minimal training instances (Figure \ref{fig:num-instances}). However, performance gains started to converge as the classification model complexity increases or additional training instances were provided. For all ms-MLP configurations, inclusion of metadata never degraded performance, however, the same was not true with CNNs where careful network configuration was required to avoid performance degradation with the addition of metadata. In practice, such metadata information is available at no extra costs, in terms of both data acquisition and computation. However, it’s utility needs to be first tested over a held out validation set. If constrained with computational resources or time, and hence restricted to simpler classifiers, it can be advantageous to include such information to help obtain improved segmentation performance. Whereas, if a state-of-the-art CNN approach is employed, the potentially negligible performance gain from metadata might not warrant its inclusion. 

Fruit detection and yield estimation performance (as distinct from pixel-wise performance) was best when combined with the segmentation results from the state-of-the-art CNN architecture. The inclusion of metadata on the ms-MLP architecture resulted in minor detection, fruit counting and yield estimation improvements. For fruit detection (from pixels to whole fruit), the WS algorithm performed better than CHT over all evaluation measures. This could be because the segmented fruits do not appear as circular objects due to noise in the segmentation output or due to heavy occlusions from non-fruit objects. Practically, the WS algorithm was also easier to tune, relying on a single hyperparameter whereas the CHT algorithm was governed by six hyperparameters.  

\subsection{Image Processing Errors}
\begin{figure}[!h]
	\centering
		\includegraphics[width=0.95\textwidth]{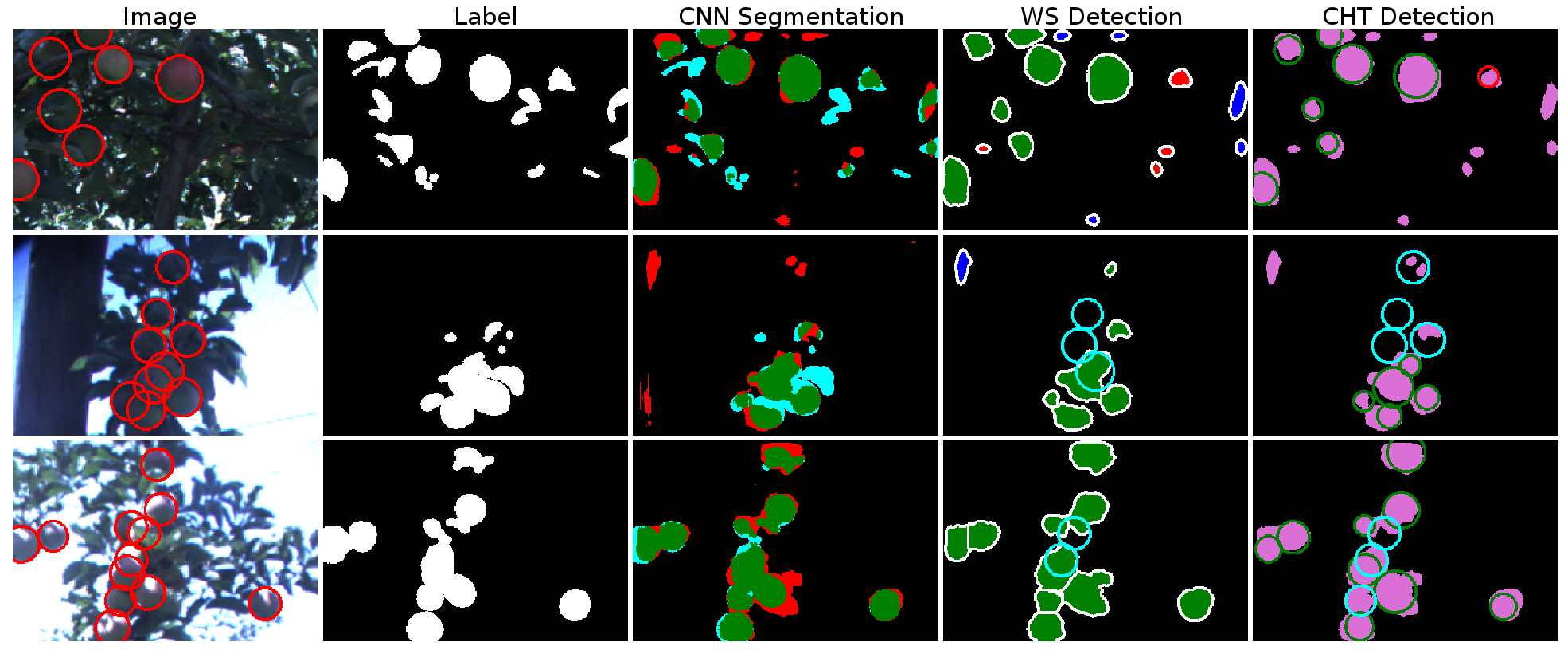}
		\caption{Instances of erroneous fruit segmentation, detection and manual labelling. The first column contains the input image data with annotated fruit detections. The second column contains the separately annotated pixel level labels, which were used to train the segmentation networks. The output from the trained CNN* network is shown in column three. The last two columns illustrate the output from the WS and CHT detection algorithms. The processed results are shown as, green: true positive, red: false positive and cyan: false negative. The examples in the first two rows illustrate discrepancies in the ground truth, between the annotated detection and the pixel-level labels. The second and third examples illustrate under counting during detection due to clustered fruit regions.}
	\label{fig:error-examples}
\end{figure}

Most common image segmentation errors were in regions with poor image quality, due to adverse illumination conditions (e.g. sun-flares and under-exposure in shadows) or in ambiguous image regions (e.g. occluded apples between green leaves). Additionally, there were errors and inconsistencies within the human labelled data, where fruits were either missed, or background fruits (in an adjacent row) were inconsistently labelled, limiting the performance and evaluation measures of the segmentation and detection algorithms. The first two rows in Figure \ref{fig:error-examples} are examples of discrepancies in the ground truth, seen through the differences in the pixel-wise and fruit-wise labels. 

The primary causes of detection errors were poor image segmentation, under counting of fruits appearing in clusters and double counting disjoint fruit regions. The WS and CHT detection algorithms are efficient at splitting fruit regions occurring in clusters of two or three apples, but are unable to discern individual fruits in larger clusters. Examples on rows two and three in Figure \ref{fig:error-examples} illustrate cases where fruit clusters are under counted by the detection algorithms. However, such large cluster occurrences were rare due to standard thinning operations employed in this orchard block, which are commonly used on orchards to optimise the quality of fruit \citep{Wang2013}. Errors with the CHT detection algorithm were also observed for instances where occluded fruit were not circular in the image. Although outside the scope of this paper, the detection results could be further improved with more specific shape based detection approaches like the ones used in \citep{Linker2012}, where fruits are defined as arc segments rather than whole circles.  

Errors in yield estimation can be attributed to erroneous fruit detections (leading to over counting or under counting), detection of fruits from background rows, and double counting or missed detections due to lack of fruit registration. In rows with sparse foliage, more of the background trees can be detected therefore skewing the yield counts. Background removal via depth estimation (e.g. through stereo vision) would be required to minimize this error. To accurately accumulate fruit counts between frames, multiple viewpoint fruit registration would be required as done in \cite{Wang2013} and \cite{Moonrinta2010}, which use stereo camera configurations to localise each fruit in $3D$ space.

\subsection{Lessons Learned}
\label{sec:lessons-learned}

The image segmentation, fruit detection and yield estimation approaches presented in this paper can be extended to other orchards, with datasets captured under variable illumination conditions over different seasons. The learning approaches for image segmentation make them flexible and adaptable to a wide range of classification problems, while the same detection framework could be used for a variety of circular fruits such as peaches, citrus, mangoes etc. For practical deployment of such a system in commercial orchards, we reflect on a number of lessons learned from this work. 

Data collected under natural illumination conditions is subject to extensive intra-class variations. Therefore, we become reliant on extensive training examples and more complex classification architectures to learn the highly non-linear distribution. However, where possible, it would be advantageous to minimise such variations by capturing image data on an overcast day or when the sun is directly overhead or in the background (e.g. in the morning or the afternoon with the camera facing the other way). A more controlled operation could involve night time imagery with strobes as done in \cite{Payne2014}, however, as mentioned in Section \ref{sec:relatedwork}, daytime operations are more desirable for integration with the current farming practices.  

Supervised image segmentation requires training examples, provided as pixel-wise labels in this paper, denoting fruit and non-fruit regions (the same as the ones used in \cite{Hung2015} and \cite{Bargoti2016ICRA}). The labelled dataset was small compared to the collection of images acquired at the orchard block, however, pixel-wise annotation was still a time consuming operation, taking approximately $1.5$ hours per $100$ images ($308\times202$). In practice, it was also challenging to manually label images consistently, particularly where adverse illumination conditions caused partial under or over exposure (e.g. some of the images in Figure \ref{fig:sub-tree-images-height}). Further work should therefore prioritise on methods that can improve labelling consistency and reduce labelling costs. 

When extending the image segmentation framework to a new dataset, a data scientist would be required to manually label a small segment of the data. The computational constraints would then drive the choice of the segmentation architecture. These costs could be minimised by using/transferring segmentation models learned over previous datasets, however, further work is required to understand the limitations. 

Differentiating from other works presented in Section \ref{sec:relatedwork}, our aim had been to minimise hand engineering features for image processing. In practice, while the neural network architectures enable this for image segmentation, the process of tuning hyperparameters is not completely trivial and is required to achieve good performance from these algorithms. Practically, it was easier to tune hyperparameters for the ms-MLP architecture compared to the CNN architecture due to its considerably shorter training time. The subsequent fruit detection and yield mapping currently rely on hand-engineered detection algorithms, however, they do not require any additional training data or feature learning. 

Due to occlusions and clustering, it is not possible to have all the apples clearly visible in the image data. Therefore, for yield estimation or prediction, the algorithm output for apple counting needs to be calibrated against a real ground truth source. This can be done at the tree, multi-tree or row-segment, row (as done in this paper) or at an orchard block level. The smaller scale data collection could overlap with sparse manual sampling operations conducted by agronomists, whereas large scale calibration could be done post-harvest from the grading and counting machines. Both operations are already performed, so the data is available at no extra cost. Further work is required to understand the amount of calibration data required, the cost of collecting it and sensitivity to collection errors. 

Finally, the system currently operates off-line, providing the grower or an agronomist with a dense spatial yield map within a day or two of obtaining the data. However, with a prediction time of $\sim 7s$ for a $1232\times1616$ image, a pre-trained detection pipeline could potentially be deployed on a robotic harvester or sprayer for real-time operation. 

\section{Conclusion}
\label{sec:conclusion}
This paper presented an image processing framework for detecting fruit and estimating yield in an orchard. General purpose feature learning algorithms were utilised for image segmentation, followed by detection and counting of individual fruits. The image segmentation was performed using a multi-scale Multi-Layered Perceptron (ms-MLP) and a Convolutional Neural Network (CNN). Within these architectures we incorporated metadata (contextual information about how the image data was captured) to explicitly capture relationships between meta-parameters and the object classes to be learnt. The pixel-wise image segmentation output was post-processed using the Watershed Segmentation (WS) and Circular Hough Transform (CHT) algorithms to detect and count individual fruits. 

The different stages of the pipeline were evaluated on image data captured over a $0.5$ ha apple orchard block using a monocular camera mounted on an unmanned ground vehicle. Captured under natural lighting conditions, the dataset covered different apple varieties with appearances ranging from red, pink-green to green. Metadata incorporated into the architectures included pixel position, row numbering and sun position relative to the camera. Following our previous work in \cite{Bargoti2016ICRA}, the metadata yielded a boost in performance with the ms-MLP network. However, the best segmentation F1-score of $0.791$ was obtained using a CNN network, with which the inclusion of metadata had negligible impact. The improved segmentation performance with the CNN also translated to more accurate fruit detection with both detection algorithms, with the best detection F1-score of 0.858 achieved using WS detection. The fruit counts were accumulated over individual rows at the orchard and compared against the post-harvest counts, where the best yield estimation fit was obtained with the same configuration, resulting in a squared correlation coefficient of $r^2=0.826$. 

The feature learning based state-of-the-art image segmentation architectures enable us to process challenging orchard image data without needing to hand engineer the feature extraction phase. To better understand their generalisation capabilities, future work will involve image segmentation and fruit detection for different fruits/orchards. For commercial realisation, future work will also investigate different labelling strategies for training the segmentation/detection algorithms, and transfer learning between different fruits/datasets.

\subsubsection*{Acknowledgments}
This work is supported by the Australian Centre for Field Robotics at The University of Sydney and Horticulture Australia Limited through project AH11009 Autonomous Perception System for Horticulture Tree Crops. Thanks to Calvin Hung for his guidance with the MLP architecture and to Kevin Sander for his support during the field trials at his apple orchard. Further information and videos available at: \url{http://sydney.edu.au/acfr/agriculture}

\bibliographystyle{apalike}
\bibliography{references}

\begin{thebibliography}{}

\bibitem[Annamalai et~al., 2004]{Annamalai2004}
Annamalai, P., Lee, W.~S., and Burks, T.~F. (2004).
\newblock {Color vision system for estimating citrus yield in real-time}.
\newblock In {\em ASAE Annual International Meeting}.

\bibitem[Atherton and Kerbyson, 1999]{Atherton1999}
Atherton, T.~J. and Kerbyson, D.~J. (1999).
\newblock {Size invariant circle detection}.
\newblock {\em Image and Vision Computing}, 17(11):795--803.

\bibitem[Bargoti and Underwood, 2015]{Bargoti2015IROS}
Bargoti, S. and Underwood, J. (2015).
\newblock {Utilising Metadata to Aid Image Classification in Orchards}.
\newblock In {\em IEEE International Conference on Intelligent Robots and
  Systems (IROS), Workshop on Alternative Sensing for Robot Perception
  (WASRoP)}.

\bibitem[Bargoti and Underwood, 2016]{Bargoti2016ICRA}
Bargoti, S. and Underwood, J. (2016).
\newblock {Image Classification with Orchard Metadata}.
\newblock In {\em IEEE International Conference on Robotics and Automation
  (ICRA) (Accepted)}.

\bibitem[Bargoti et~al., 2015]{Bargoti:2015}
Bargoti, S., Underwood, J.~P., Nieto, J.~I., and Sukkarieh, S. (2015).
\newblock {A Pipeline for Trunk Detection in Trellis Structured Apple
  Orchards}.
\newblock {\em Journal of Field Robotics}, 32(8):1075--1094.

\bibitem[Brown and Susstrunk, 2011]{Brown2011}
Brown, M. and Susstrunk, S. (2011).
\newblock {Multi-spectral SIFT for scene category recognition}.
\newblock In {\em IEEE Conference on Computer Vision and Pattern Recognition
  (CVPR)}, pages 177--184.

\bibitem[Brust et~al., 2015]{Brust15:CPN}
Brust, C.-a., Sickert, S., Simon, M., Rodner, E., and Denzler, J. (2015).
\newblock {Convolutional Patch Networks with Spatial Prior for Road Detection
  and Urban Scene Understanding}.
\newblock In {\em International Conference on Computer Vision Theory and
  Applications (VISAPP)}, pages 510--517.

\bibitem[Chhabra et~al., 2012]{Chhabra2012}
Chhabra, M., Gupta, A., Mehrotra, P., and Reel, S. (2012).
\newblock {Automated Detection of Fully and Partially Riped Mango by Machine
  Vision}.
\newblock In {\em Proceedings of the International Conference on Soft Computing
  for Problem Solving (SocProS)}, volume 131 of {\em Advances in Intelligent
  and Soft Computing}, pages 153--164. Springer India.

\bibitem[Ciresan et~al., 2012]{Ciresan2012}
Ciresan, D., Giusti, A., Gambardella, L.~M., and Schmidhuber, J. (2012).
\newblock {Deep Neural Networks Segment Neuronal Membranes in Electron
  Microscopy Images}.
\newblock In {\em Advances in Neural Information Processing Systems 25}, pages
  2843--2851. Curran Associates, Inc.

\bibitem[Erhan et~al., 2010]{Erhan:2010}
Erhan, D., Bengio, Y., Courville, A., Manzagol, P.-A., Vincent, P., and Bengio,
  S. (2010).
\newblock {Why Does Unsupervised Pre-training Help Deep Learning?}
\newblock {\em The Journal of Machine Learning Research}, 11:625--660.

\bibitem[Farabet et~al., 2013]{Farabet2013}
Farabet, C., Couprie, C., Najman, L., and LeCun, Y. (2013).
\newblock {Learning Hierarchical Features for Scene Labeling}.
\newblock {\em IEEE Transactions on Pattern Analysis and Machine Intelligence},
  35(8):1915--1929.

\bibitem[Font et~al., 2015]{Font2015}
Font, D., Tresanchez, M., Mart{\'{\i}}nez, D., Moreno, J., Clotet, E., and
  Palac{\'{\i}}n, J. (2015).
\newblock {Vineyard Yield Estimation Based on the Analysis of High Resolution
  Images Obtained with Artificial Illumination at Night}.
\newblock {\em Sensors}, 15(4):8284.

\bibitem[Ganin and Lempitsky, 2014]{GaninL14}
Ganin, Y. and Lempitsky, V.~S. (2014).
\newblock {{$N^4$}-Fields: Neural Network Nearest Neighbor Fields for Image
  Transforms}.
\newblock {\em CoRR}, abs/1406.6.

\bibitem[Goodfellow and Warde-Farley, 2013]{pylearn2_arxiv_2013}
Goodfellow, I. and Warde-Farley, D. (2013).
\newblock {Pylearn2: a machine learning research library}.
\newblock {\em arXiv preprint arXiv:1308.4214}, pages 1--9.

\bibitem[Hung et~al., 2013]{Hung2013}
Hung, C., Nieto, J., Taylor, Z., Underwood, J., and Sukkarieh, S. (2013).
\newblock {Orchard fruit segmentation using multi-spectral feature learning}.
\newblock In {\em IEEE International Conference on Intelligent Robots and
  Systems (IROS)}, pages 5314--5320.

\bibitem[Hung et~al., 2015]{Hung2015}
Hung, C., Underwood, J., Nieto, J., and Sukkarieh, S. (2015).
\newblock {A Feature Learning Based Approach for Automated Fruit Yield
  Estimation}.
\newblock In {\em Field and Service Robotics (FSR)}, pages 485--498. Springer.

\bibitem[Ji et~al., 2012]{Ji2012}
Ji, W., Zhao, D., Cheng, F., Xu, B., Zhang, Y., and Wang, J. (2012).
\newblock {Automatic recognition vision system guided for apple harvesting
  robot}.
\newblock {\em Computers {\&} Electrical Engineering}, 38(5):1186--1195.

\bibitem[Jimenez et~al., 2000]{Jimenez2000a}
Jimenez, A.~R., Ceres, R., and Pons, J.~L. (2000).
\newblock {A survey of computer vision methods for locating fruit on trees}.
\newblock {\em Transactions of the ASAE-American Society of Agricultural
  Engineers}, 43(6):1911--1920.

\bibitem[Kapach et~al., 2012]{Kapach:2012}
Kapach, K., Barnea, E., Mairon, R., Edan, Y., and Ben-Shahar, O. (2012).
\newblock {Computer Vision for Fruit Harvesting Robots - state of the Art and
  Challenges Ahead}.
\newblock {\em International Journal of Computational Vision and Robotics},
  3(1-2):4--34.

\bibitem[Kim et~al., 2015]{Kim2015}
Kim, D., Choi, H., Choi, J., Yoo, S.~J., and Han, D. (2015).
\newblock {A Novel Red Apple Detection Algorithm Based on AdaBoost Learning}.
\newblock {\em IEIE Transactions on Smart Processing {\&} Computing},
  4(4):265--271.

\bibitem[Krizhevsky et~al., 2012]{Krizhevsky2012}
Krizhevsky, A., Sutskever, I., and Hinton, G.~E. (2012).
\newblock {Imagenet classification with deep convolutional neural networks}.
\newblock In {\em Advances in neural information processing systems}, pages
  1097--1105.

\bibitem[Kurtulmus et~al., 2014]{Kurtulmus2014}
Kurtulmus, F., Lee, W., and Vardar, A. (2014).
\newblock {Immature peach detection in colour images acquired in natural
  illumination conditions using statistical classifiers and neural network}.
\newblock {\em Precision Agriculture}, 15(1):57--79.

\bibitem[Li et~al., 2011]{Li2011}
Li, P., Lee, S.-h., and Hsu, H.-Y. (2011).
\newblock {Study on citrus fruit image data separability by segmentation
  methods}.
\newblock {\em Procedia Engineering}, 23:408--416.

\bibitem[Linker et~al., 2012]{Linker2012}
Linker, R., Cohen, O., and Naor, A. (2012).
\newblock {Determination of the number of green apples in RGB images recorded
  in orchards}.
\newblock {\em Computers and Electronics in Agriculture}, 81:45--57.

\bibitem[Liu et~al., 2015]{Liu2015}
Liu, S., Whitty, M., and Cossell, S. (2015).
\newblock {Automatic grape bunch detection in vineyards for precise yield
  estimation}.

\bibitem[Long et~al., 2015]{Long2015}
Long, J., Shelhamer, E., and Darrell, T. (2015).
\newblock {Fully Convolutional Networks for Semantic Segmentation}.
\newblock {\em Proceedings of the IEEE Conference on Computer Vision and
  Pattern Recognition (CVPR)}, pages 3431--3440.

\bibitem[Martens, 2010]{Martens2010}
Martens, J. (2010).
\newblock {Deep learning via Hessian-free optimization}.
\newblock In {\em Proceedings of the 27th International Conference on Machine
  Learning (ICML)}, pages 735--742.

\bibitem[Moonrinta et~al., 2010]{Moonrinta2010}
Moonrinta, J., Chaivivatrakul, S., Dailey, M.~N., and Ekpanyapong, M. (2010).
\newblock {Fruit detection, tracking, and 3D reconstruction for crop mapping
  and yield estimation}.
\newblock In {\em International Conference on Control Automation Robotics {\&}
  Vision (ICARCV)}, pages 1181--1186.

\bibitem[Ning et~al., 2005]{Ning2005}
Ning, F., Delhomme, D., LeCun, Y., Piano, F., Bottou, L., and Barbano, P.~E.
  (2005).
\newblock {Toward automatic phenotyping of developing embryos from videos}.
\newblock {\em IEEE Transactions on Image Processing}, 14(9):1360--1371.

\bibitem[Nuske et~al., 2014]{Nuske2014}
Nuske, S., Wilshusen, K., Achar, S., Yoder, L., and Singh, S. (2014).
\newblock {Automated visual yield estimation in vineyards}.
\newblock {\em Journal of Field Robotics}, 31(5):837--860.

\bibitem[Payne and Walsh, 2014]{Payne2014Book}
Payne, A. and Walsh, K. (2014).
\newblock {Machine vision in estimation of fruit crop yield}.
\newblock In {\em Plant Image Analysis: Fundamentals and Applications},
  chapter~16, pages 329--374. CRC Press.

\bibitem[Payne et~al., 2014]{Payne2014}
Payne, a., Walsh, K., Subedi, P., and Jarvis, D. (2014).
\newblock {Estimating mango crop yield using image analysis using fruit at
  'stone hardening' stage and night time imaging}.
\newblock {\em Computers and Electronics in Agriculture}, 100:160--167.

\bibitem[Pinheiro and Collobert, 2013]{PinheiroC13}
Pinheiro, P. H.~O. and Collobert, R. (2013).
\newblock {Recurrent Convolutional Neural Networks for Scene Parsing}.
\newblock {\em CoRR}, abs/1306.2.

\bibitem[Qiang et~al., 2014]{Qiang2014}
Qiang, L., Jianrong, C., Bin, L., Lie, D., and Yajing, Z. (2014).
\newblock {Identification of fruit and branch in natural scenes for citrus
  harvesting robot using machine vision and support vector machine}.
\newblock {\em International Journal of Agricultural and Biological
  Engineering}, 7(2):115--121.

\bibitem[Rao et~al., 2014]{Rao2014}
Rao, D., {De Deuge}, M., Nourani-Vatani, N., Douillard, B., Williams, S.~B.,
  and Pizarro, O. (2014).
\newblock {Multimodal learning for autonomous underwater vehicles from visual
  and bathymetric data}.
\newblock In {\em IEEE International Conference on Robotics and Automation
  (ICRA)}, pages 3819--3825.

\bibitem[Regunathan and Lee, 2005]{Regunathan2005}
Regunathan, M. and Lee, W.~S. (2005).
\newblock {Citrus fruit identification and size determination using machine
  vision and ultrasonic sensors}.
\newblock In {\em ASAE annual international meeting}.

\bibitem[Roerdink and Meijster, 2000]{Roerdink2000}
Roerdink, J. B. T.~M. and Meijster, A. (2000).
\newblock {The watershed transform: Definitions, algorithms and parallelization
  strategies}.
\newblock {\em Fundamenta informaticae}, 41(1-2):187--228.

\bibitem[Sa et~al., 2015]{Sa2015}
Sa, I., McCool, C., Lehnert, C., and Perez, T. (2015).
\newblock {On Visual Detection of Highly-occluded Objects for Harvesting
  Automation in Horticulture}.
\newblock In {\em IEEE International Conference on Robotics and Automation
  (ICRA)}. ICRA.

\bibitem[Sengupta and Lee, 2014]{Sengupta2014}
Sengupta, S. and Lee, W.~S. (2014).
\newblock {Identification and determination of the number of immature green
  citrus fruit in a canopy under different ambient light conditions}.
\newblock {\em Biosystems Engineering}, 117:51--61.

\bibitem[Sermanet et~al., 2013]{sermanet2013overfeat}
Sermanet, P., Eigen, D., Zhang, X., Mathieu, M., Fergus, R., and LeCun, Y.
  (2013).
\newblock {Overfeat: Integrated recognition, localization and detection using
  convolutional networks}.
\newblock {\em arXiv preprint arXiv:1312.6229}.

\bibitem[Silwal et~al., 2014]{Silwal2014}
Silwal, A., Gongal, A., and Karkee, M. (2014).
\newblock {Identification of red apples in field environment with over the row
  machine vision system}.
\newblock {\em Agricultural Engineering International: CIGR Journal},
  16(4):66--75.

\bibitem[Song et~al., 2014]{Song2014}
Song, Y., Glasbey, C.~a., Horgan, G.~W., Polder, G., Dieleman, J.~a., and
  van~der Heijden, G. W. a.~M. (2014).
\newblock {Automatic fruit recognition and counting from multiple images}.
\newblock {\em Biosystems Engineering}, 118(1):203--215.

\bibitem[Stajnko et~al., 2009]{Stajnko2009}
Stajnko, D., Rakun, J., and Blanke, M.~M. (2009).
\newblock {Modelling apple fruit yield using image analysis for fruit colour,
  shape and texture}.
\newblock {\em European Journal of Horticultural Science}, 74(6):260--267.

\bibitem[Tighe and Lazebnik, 2013]{tighe2013superparsing}
Tighe, J. and Lazebnik, S. (2013).
\newblock {Superparsing}.
\newblock {\em International Journal of Computer Vision}, 101(2):329--349.

\bibitem[Wang et~al., 2013]{Wang2013}
Wang, Q., Nuske, S., Bergerman, M., and Singh, S. (2013).
\newblock {Automated Crop Yield Estimation for Apple Orchards}.
\newblock In {\em Experimental Robotics}, volume~88 of {\em Springer Tracts in
  Advanced Robotics}, pages 745--758. Springer International Publishing.

\bibitem[Wijethunga et~al., 2009]{Wijethunga2009}
Wijethunga, P., Samarasinghe, S., Kulasiri, D., and Woodhead, I. (2009).
\newblock {Towards a generalized colour image segmentation for kiwifruit
  detection}.
\newblock In {\em International Conference Image and Vision Computing New
  Zealand (IVCNZ)}, pages 62--66.

\bibitem[Yamamoto et~al., 2014]{Yamamoto2014}
Yamamoto, K., Guo, W., Yoshioka, Y., and Ninomiya, S. (2014).
\newblock {On Plant Detection of Intact Tomato Fruits Using Image Analysis and
  Machine Learning Methods}.
\newblock {\em Sensors}, 14(7):12191--12206.

\end{thebibliography}

\end{document}